%% file: main.tex
\PassOptionsToPackage{table}{xcolor}

\documentclass[10pt,twocolumn,letterpaper]{article}
\usepackage{tikz}
\usetikzlibrary{positioning, shapes, arrows.meta}
\usepackage{bm}
\usepackage{graphicx}
\usepackage{makecell}
\usepackage{amsmath}
\usepackage{float}
\usepackage[table]{xcolor}
\definecolor{mygreen}{RGB}{0,150,0}
\definecolor{mylightgreen}{RGB}{200,255,200}

\usepackage{hhline}
\usepackage{geometry}
\usepackage{booktabs}
\usepackage{arydshln}
\usepackage{amsfonts}
\usepackage{multirow}
\usepackage{subcaption}
\usepackage{mathtools}

\usepackage{amsthm}
\newtheorem{theorem}{Theorem}
\usepackage{algorithm}      
\usepackage{algpseudocode}  
\newtheorem{lemma}{Lemma}
\usepackage{multirow}
\usepackage{xcolor}

\usepackage{amsmath,amssymb}
\usepackage{adjustbox}

\usepackage[table]{xcolor}
\definecolor{teaserblue}{RGB}{235, 180, 123}
\usepackage{xcolor}
\definecolor{ForestGreen}{RGB}{34,139,34}
\usepackage{pifont}

\newcommand\blfootnote[1]{%
  \begingroup
  \renewcommand\thefootnote{}\footnote{#1}%
  \addtocounter{footnote}{-1}%
  \endgroup
}

\usepackage{cvpr}
\newtheorem{proposition}{Proposition}

\newcommand{\cmark}{\checkmark}
\newcommand{\xmark}{\ding{55}}
\definecolor{cvprblue}{rgb}{0.21,0.49,0.74}

\usepackage[pagebackref,breaklinks,colorlinks,
            citecolor=cvprblue,
            linkcolor=cvprblue,
            urlcolor=cvprblue]{hyperref}
\renewcommand*{\backref}[1]{}
\renewcommand*{\backrefalt}[4]{%
  \ifcase #1 \or {\color{cvprblue}#2}\else {\color{cvprblue}#2}\fi}

\def\paperID{42148} 
\def\confName{CVPR}
\def\confYear{2026}

\usetikzlibrary{decorations.pathreplacing}

\title{ELSA: Exact Linear-Scan Attention for Fast and Memory-Light \\Vision Transformers}

\author{
    Chih-Chung Hsu \quad
    Xin-Di Ma \quad Wo-Ting Liao \quad  Chia-Ming Lee \\
{Advanced Computer Vision Laboratory, National Yang Ming Chiao Tung University} }

\begin{document}

\twocolumn[{%
\renewcommand\twocolumn[1][]{#1}%
\maketitle
\begin{center}
\centering
\captionsetup{type=figure}
\end{center}
\vspace{-6mm}
}]

\maketitle
\input{sec/1-abstract}
\blfootnote{This study was supported in part by the National Science and Technology Council (NSTC), Taiwan, under grants 112-2221-E-006-157-MY3, 114-2627-M-A49-003, and 114-2218-E-035-001. We thank the National Center for NCHC of NARLabs in Taiwan for providing computational and storage resources. Corresponding author: Chih-Chung Hsu.}
\input{sec/2-introduction}
\input{sec/3-relatedwork}
\input{sec/4-method}

\input{sec/5-experiment}
\input{sec/6-conclusion}


{
    \small
    \bibliographystyle{ieeenat_fullname}
\bibliography{main}

\input{sec/7-suppl}  

}


\end{document}

%% file: sec/1-abstract.tex
\vspace{-2mm}\begin{abstract}
Existing attention accelerators often trade exact softmax semantics,
depend on fused Tensor Core kernels, or incur sequential depth that limits FP32 throughput on long sequences. We present \textbf{ELSA}, an algorithmic reformulation of online softmax attention that (i)~preserves exact softmax semantics in real arithmetic with a \emph{provable} $\mathcal{O}(u\log n)$ FP32 relative error bound;
(ii)~casts the online softmax update as a prefix scan over an associative monoid $(m,S,W)$, yielding $O(n)$ extra memory and $O(\log n)$ parallel depth; and (iii)~is Tensor-Core independent, implemented in Triton and CUDA C++, and deployable as a \emph{drop-in replacement} requiring no retraining or weight modification.
Unlike FlashAttention-2/3, which rely on HMMA/GMMA Tensor Core instructions and provide no compatible FP32 path, ELSA operates identically on A100s and resource-constrained edge devices such as
Jetson TX2---making it the only hardware-agnostic exact-attention kernel that reduces parallel depth to $O(\log n)$ at full precision.
On A100 FP32 benchmarks (1K--16K tokens), ELSA delivers $1.3$--$3.5\times$ speedup over memory-efficient SDPA and $1.97$--$2.27\times$ on BERT; on Jetson TX2, ELSA achieves $1.5$--$1.6\times$ over Math (64--900 tokens), with $17.8$--$20.2\%$ throughput gains under LLaMA-13B offloading at $\ge$32K. In FP16, ELSA approaches hardware-fused baselines at long sequences while retaining full FP32 capability, offering a unified kernel for
high-precision inference across platforms. Our code and implementation are available at \url{https://github.com/ming053l/ELSA}.
\end{abstract}
\vspace{-7mm}

%% file: sec/2-introduction.tex
\section{Introduction}

Vision Transformers (ViT)~\cite{vaswani2017,dosovitskiy2021} have achieved
remarkable success across computer vision, but their Multi-Head
Self-Attention (MHSA) suffers from quadratic memory scaling.
A 4K image ($4096\times4096$), tokenized at a $4{\times}4$
patch size typical of high-resolution scientific models, yields
${\approx}1\text{M}$ tokens and requires over 4~TB of GPU
memory to materialize the full attention matrix in FP32---far exceeding
commercial hardware capacity.
This $\mathcal{O}(n^2)$ bottleneck is particularly critical in
high-precision domains such as medical imaging, hyperspectral analysis, and
super-resolution microscopy, where FP32 accuracy is
non-negotiable~\cite{fan2024vitarvisiontransformerresolution,zhang2021multiscale,%
yao2024hirivitscalingvisiontransformer,leroy2024winwin,dehghani2023patch}.
Accelerating attention while preserving exactness and precision is thus
essential for scaling ViTs to high-resolution applications.

\textbf{Challenges.}
Existing approaches remain limited for high-speed, high-precision inference.
Current strategies fall into three paradigms:
\emph{approximate attention}~\cite{Performer,Linformer,Zaheer2020}, which
linearize complexity via low-rank projections or sparse patterns;
\emph{hardware-fused kernels} like
FlashAttention-2/3~\cite{dao2023,shah2024flashattention}, which achieve
linear memory through block-tiling and Tensor Core instructions; and
\emph{memory-efficient implementations} such as
ME-SDPA~\cite{xformers2022}, which reduce memory via sequential computation.
Each faces critical limitations: approximate methods alter the attention
operator, requiring retraining that is infeasible for foundation models trained over
millions of GPU hours and cost-intensive data curation (e.g., CLIP~\cite{radford2021},
SAM~\cite{kirillov2023}, LLaMA~\cite{touvron2023llama}, and 3D Large Reconstruction Model like VGGT~\cite{wang2025vggt}); FlashAttention's FP32 path reverts to
untuned SIMD without fused HMMA kernels and requires per-generation tuning;
ME-SDPA, despite supporting FP32, suffers $\mathcal{O}(n/T_k)$ sequential
depth---\textcolor{cvprblue}{\textbf{$3.5\times$ slower than ELSA at 16K
tokens}}.
We therefore restrict comparisons to \emph{exact} drop-in implementations
(Math kernel, ME-SDPA, FA2/FA3); approximate methods fall outside fair
evaluation scope as they require retraining every benchmark model.

\textbf{Problem Formulation.}
\emph{What remains missing is an exact, inherently parallel reformulation of
attention that requires no specialized hardware.}
The online softmax recurrence of Rabe and Staats~\cite{rabe2021} reduces
memory to $\mathcal{O}(n)$ but remains inherently sequential. For the
${\approx}1\text{M}$-token 4K example above, a GPU must synchronize over
one million steps, leaving pipelines idle.
Breaking this sequential bottleneck is the key challenge for efficient FP32
attention.

To address this, we propose \textbf{Exact Linear-Scan Attention (ELSA)},
which reformulates the online softmax recurrence as a prefix scan over an
associative monoid, achieving
\textbf{$\mathcal{O}(\log n)$ parallel depth} and
\textbf{$\mathcal{O}(n)$ memory} with
\underline{exact FP32 semantics}, \underline{no retraining}, and
\underline{no Tensor Core dependency}
(in FP32 we provide a provable bound under Assumption~1).
Table~\ref{tab:attention_tradeoff} summarizes how ELSA uniquely combines
exactness, FP32 efficiency, and GPU-agnostic design.
Our contributions are as follows:

\begin{itemize}

\item \textbf{Provable FP32 error bound and deployable kernel.}
We formalize the online softmax merge operator $\oplus$ on state triples
$(m,S,W)\!\in\!\bar{\mathbb{R}}\times\mathbb{R}_{\ge 0}\times\mathbb{R}^{d_v}$ as a
monoid---going beyond prior theoretical treatments~\cite{feng2024} by
proving closure, associativity, and identity---and derive a
\emph{kernel-aware} $\mathcal{O}(u\log n)$\footnote{We write $u$ for FP32
unit roundoff; FP64 would be denoted by $u_{64}$.} FP32 relative error
bound tied to scan merge depth, enabling provable numerical stability for
million-token sequences on real hardware.

\item \textbf{Hardware-agnostic implementation.}
We implement a two-level scan (intra-block Hillis--Steele, inter-block
Blelloch) in Triton and CUDA C++ without Tensor Core dependencies, achieving
high-throughput FP32 inference across platforms from data-centre GPUs
(e.g., A100, L4) to embedded devices (Jetson TX2).

\item \textbf{Retraining-free deployment.}
ELSA preserves exact softmax semantics as a drop-in replacement for
pretrained large foundation models, such as CLIP, ViT, LLaMA, and VGGT, enabling immediate acceleration without
fine-tuning, which is critical for foundation models trained over millions of GPU
hours.
\item \textbf{Comprehensive evaluation.}
Across vision and language benchmarks,
\textbf{(i)}~in FP32 ELSA improves latency over
memory-efficient SDPA by $1.3$--$3.5\times$ while using comparable peak
memory; \textbf{(ii)}~in FP16 ELSA approaches
FlashAttention-2/3 throughput at long sequences while achieving the lowest
peak memory among exact kernels (e.g., $0.19$\,GB vs.\
$0.29$/$0.36$\,GB at 16K, Figure \ref{fig:table1}). \textbf{(iii)}~under host-device offloading, ELSA delivers {{17.8--20.2\%}} throughput gains over
  ME-SDPA at $\ge$32K tokens (LLaMA-13B, FP32).

\end{itemize}

\begin{table}[t]
\captionsetup{aboveskip=2pt,belowskip=0pt}
\centering
\caption{\textbf{Method Comparison.} ELSA is exact, hardware-agnostic,
and retrain-free---a drop-in replacement requiring no architecture-specific
primitives---while achieving $O(\log n)$ depth and $O(n)$ memory in FP32.}
\label{tab:unified_attn_landscape_compact}
\setlength{\tabcolsep}{3pt}
\renewcommand{\arraystretch}{1.05}
\scriptsize
\resizebox{\columnwidth}{!}{
\begin{tabular}{lcccccc}
\toprule
\textbf{Method} & Exact & FP32 & \makecell{HW-\\agnostic} & \makecell{Retrain-\\free} & Depth (per head) & Extra mem. \\
\midrule
Standard SDPA              & \cmark & \xmark & \cmark & \cmark & $O(n)$        & $O(n^2)$   \\
ME\mbox{-}SDPA (xFormers)  & \cmark & \cmark & \cmark & \cmark & $O(n/T_k)$    & $O(T_k d)$ \\
FlashAttn\mbox{-}2/3       & \cmark & \xmark & \xmark & \cmark & $O(n/T_k)$    & $O(T_k d)$ \\
Linear Attention           & \xmark & \cmark & \cmark & \xmark & $O(\log n)^{\dagger}$ & $O(n)$ \\
\rowcolor{gray!12}
\textbf{ELSA (Ours)}       & \cmark & \cmark & \cmark & \cmark & $\bm{O(\log n)}$ & $O(n)$ \\
\bottomrule
\end{tabular}}\label{tab:attention_tradeoff}
\vspace{2pt}
\footnotesize
\emph{Notes:} Depth for blockwise kernels scales as $O(n/T_k)$; ELSA uses a
two-level scan (Hillis--Steele\,+\,Blelloch). ``FP32'' denotes a
high-throughput FP32 path without fused Tensor Cores.
${}^{\dagger}$Depth shown for representative linear-kernel variants; exactness
is not preserved.
\vspace{-5mm}
\end{table}

%% file: sec/3-relatedwork.tex
\vspace{-3mm}
\section{Related Works}

\noindent\textbf{Attention Approximation.}
A major research branch approximates the attention matrix to achieve
$\mathcal{O}(n)$ complexity, but at the cost of modifying the attention
operator itself.
Performer~\cite{Performer} applies random feature maps to linearize the
exponential kernel, replacing exact softmax with a kernel approximation
that enables linear-time computation but introduces approximation error.
Linformer~\cite{Linformer} projects keys and values to a lower-dimensional
space via learned linear projections, reducing the effective sequence length
but losing full token-to-token interactions.
Sparse attention methods such as BigBird~\cite{Zaheer2020} and Sparse
Transformers~\cite{Child2019} restrict attention to predefined patterns
(e.g., local windows, random connections, global tokens), discarding most
pairwise interactions to achieve sparsity.
Nystr\"{o}mformer~\cite{Xiong2021} approximates the attention matrix via the
Nystr\"{o}m method using landmark tokens, and Reformer~\cite{Kitaev2020}
employs locality-sensitive hashing to cluster similar queries and keys.
More recently, DeltaNet~\cite{wang2024deltanet} parallelizes linear
Transformers with the delta rule over sequence length using a
hardware-efficient chunk-wise scan, achieving $\mathcal{O}(n)$ training
complexity while retaining recurrent expressiveness.
While these methods achieve attractive theoretical complexity, they
inevitably modify the attention operator and require training a model
from scratch to compensate for the approximation---impractical for
foundation models such as CLIP~\cite{radford2021} or
SAM~\cite{kirillov2023} trained over millions of GPU hours and
labour-intensive data curation.
In contrast, ELSA preserves \textbf{exact softmax
semantics} and serves as a \textbf{drop-in
replacement without any retraining}.

\noindent\textbf{Hardware-fused Attention.}
FlashAttention-2 (FA2)~\cite{dao2023} pioneered tiling attention to fit
on-chip memory, achieving linear memory use and $2$--$4\times$ speedups by
fusing $QK^{\top}$, softmax, and $V$ within shared memory using
$16{\times}16$ tiles aligned to Tensor Core (HMMA) instructions.
FlashAttention-3 (FA3)~\cite{shah2024flashattention} further improves
throughput via finer tile partitioning and warp-specialization.
Both kernels, however, depend critically on HMMA/GMMA instructions optimized
for FP16/BF16; without them, \textbf{execution reverts
to SIMD paths and the throughput advantage largely disappears}.
Both are also tightly coupled to specific GPU generations, complicating
deployment on older or embedded devices.

\noindent\textbf{The Case for Exact FP32 Attention.}
Despite the prevalence of BF16/FP16, FP32 remains indispensable in several
settings: \emph{hyperspectral imaging}, where 12--16-bit sensor dynamic
range is irrecoverable under half-precision
accumulation~\cite{zhong2021spectral,hong2022spectralformer};
\emph{medical image analysis}, where clinical accuracy requirements preclude
any approximation~\cite{fan2024vitarvisiontransformerresolution,leroy2024winwin};
\emph{zero-shot deployment} of foundation models
(CLIP~\cite{radford2021}, SAM~\cite{kirillov2023}, and VGGT~\cite{wang2025vggt}), where retraining to
recalibrate precision-induced drift is infeasible; and
\emph{mixed-precision pipelines} that require a numerically auditable FP32
attention sub-module.
In all these cases, FA2/FA3's FP32 path is uncompetitive and ME-SDPA's
sequential depth imposes a \textcolor{cvprblue}{\textbf{$3.5\times$ latency
penalty at 16K tokens}}.
ELSA addresses this gap by treating softmax as an \emph{algorithmic}
problem: an associative monoid over $(m,S,W)$ solved via prefix-scan,
delivering \textbf{exact, high-throughput FP32
inference} across platforms, from A100s to Jetson devices,
\textbf{without any Tensor Core dependency}.

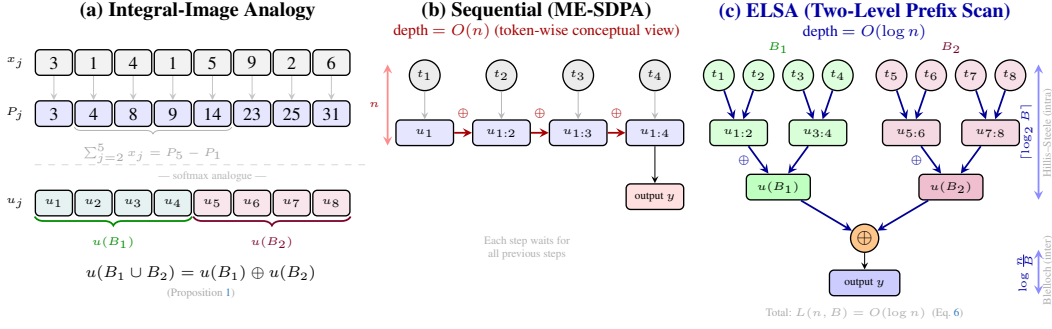
\begin{figure*}[t]
\centering
\scalebox{0.85}{
\begin{tikzpicture}[
  >=stealth, semithick,
  tok/.style={draw, circle, fill=gray!15, minimum size=0.48cm,
              font=\tiny, inner sep=0pt},
  sbox/.style={draw, rounded corners=2pt, fill=#1,
               minimum width=0.9cm, minimum height=0.4cm,
               font=\tiny, inner sep=2pt},
  sbox/.default={blue!10},
  mop/.style={draw, circle, fill=orange!45,
              minimum size=0.44cm, font=\small, inner sep=0pt},
  sarr/.style={->, thick, red!65!black},
  parr/.style={->, thick, blue!58!black},
]

\begin{scope}[xshift=0cm]
  \node[font=\small\bfseries] at (2.25,4.55) {(a) Integral-Image Analogy};

  \foreach \i/\v in {1/3,2/1,3/4,4/1,5/5,6/9,7/2,8/6}{
    \node[sbox={gray!10}, minimum width=0.58cm]
      (v\i) at ({(\i-1)*0.62},3.75) {\footnotesize\v};
  }
  \node[font=\tiny,left=1pt] at (v1.west) {$x_j$};

  \foreach \i/\v in {1/3,2/4,3/8,4/9,5/14,6/23,7/25,8/31}{
    \node[sbox={blue!10}, minimum width=0.58cm]
      (p\i) at ({(\i-1)*0.62},2.95) {\footnotesize\v};
  }
  \node[font=\tiny,left=1pt] at (p1.west) {$P_j$};
  \foreach \i in {1,...,8}{ \draw[->,thin,gray!45] (v\i)--(p\i); }

  \draw[decorate,decoration={brace,amplitude=3pt,mirror},thin,gray!60]
    (p2.south west) -- (p5.south east)
    node[midway,below=4pt,font=\tiny,gray!70]
      {$\textstyle\sum_{j=2}^{5}x_j = P_5 - P_1$};

  \draw[thin,gray!30,dashed] (-0.3,2.15)--(4.9,2.15);
  \node[font=\tiny,gray!55,align=center] at (2.48,1.97)
    {--- softmax analogue ---};

  \foreach \i in {1,...,4}{
    \node[sbox={teal!10},minimum width=0.58cm]
      (u\i) at ({(\i-1)*0.62},1.55) {\tiny$u_{\i}$};
  }
  \foreach \i in {5,...,8}{
    \node[sbox={purple!10},minimum width=0.58cm]
      (u\i) at ({(\i-1)*0.62},1.55) {\tiny$u_{\i}$};
  }
  \node[font=\tiny,left=1pt] at (u1.west) {$u_j$};

  \draw[decorate,decoration={brace,amplitude=4pt,mirror},
        green!55!black,thick]
    (u1.south west) -- (u4.south east)
    node[midway,below=5pt,font=\tiny,green!55!black]{$u(B_1)$};
  \draw[decorate,decoration={brace,amplitude=4pt,mirror},
        purple!65!black,thick]
    (u5.south west) -- (u8.south east)
    node[midway,below=5pt,font=\tiny,purple!65!black]{$u(B_2)$};

  \node[font=\scriptsize] at (2.3,0.45)
    {$u(B_1\cup B_2)=u(B_1)\oplus u(B_2)$};
  \node[font=\tiny,gray!65] at (2.3,0.13)
    {(Proposition~\ref{prop:block})};
\end{scope}

\begin{scope}[xshift=5.8cm]
  \node[font=\small\bfseries] at (1.75,4.55) {(b) Sequential (ME-SDPA)};
  \node[font=\scriptsize,red!65!black] at (1.75,4.25) {depth $=O(n)$ (token-wise conceptual view)};

    \foreach \i in {1,...,4}{
    \node[tok] (tb\i) at ({(\i-1)*1.2},3.55) {\tiny$t_\i$};
  }

  \node[sbox] (sb1) at (0.00,2.65) {\tiny$u_1$};
  \node[sbox] (sb2) at (1.20,2.65) {\tiny$u_{1:2}$};
  \node[sbox] (sb3) at (2.40,2.65) {\tiny$u_{1:3}$};
  \node[sbox] (sb4) at (3.60,2.65) {\tiny$u_{1:4}$};

  \foreach \i in {1,...,4}{ \draw[->,thin,gray!55] (tb\i)--(sb\i); }

  \draw[sarr] (sb1)--(sb2) node[midway,above=3pt,font=\tiny]{$\oplus$};
  \draw[sarr] (sb2)--(sb3) node[midway,above=3pt,font=\tiny]{$\oplus$};
  \draw[sarr] (sb3)--(sb4) node[midway,above=3pt,font=\tiny]{$\oplus$};

  \draw[<->,thick,red!45] (-0.58,2.45)--(-0.58,3.72)
    node[midway,left,font=\tiny,red!65!black]{$n$};
    
  \node[sbox={red!12}] (outb) at (3.60,1.65) {\tiny output $y$};
  \draw[->,thin] (sb4)--(outb);

  \node[font=\tiny,align=center,gray!60] at (1.6,0.85)
    {Each step waits for\\all previous steps};
\end{scope}

\begin{scope}[xshift=10.4cm]
  \node[font=\small\bfseries,blue!65!black] at (2.28,4.55)
    {(c) ELSA (Two-Level Prefix  Scan)};
  \node[font=\scriptsize,blue!65!black] at (2.28,4.22)
    {depth $=O(\log n)$};

  \foreach \i in {1,...,4}{
    \node[tok,fill=green!13] (tc\i) at ({(\i-1)*0.62},3.55) {\tiny$t_\i$};
  }
  \foreach \i in {5,...,8}{
    \node[tok,fill=purple!12] (tc\i) at ({2.7+(\i-5)*0.62},3.55)
      {\tiny$t_\i$};
  }

  \node[font=\tiny,green!55!black]  at (0.93,3.98) {$B_1$};
  \node[font=\tiny,purple!60!black] at (3.63,3.98) {$B_2$};

  \node[sbox={green!15}] (c1a) at (0.31,2.65) {\tiny$u_{1:2}$};
  \node[sbox={green!15}] (c1b) at (1.55,2.65) {\tiny$u_{3:4}$};
  \draw[parr] (tc1)--(c1a); \draw[parr] (tc2)--(c1a);
  \draw[parr] (tc3)--(c1b); \draw[parr] (tc4)--(c1b);

  \node[sbox={purple!15}] (c2a) at (3.01,2.65) {\tiny$u_{5:6}$};
  \node[sbox={purple!15}] (c2b) at (4.25,2.65) {\tiny$u_{7:8}$};
  \draw[parr] (tc5)--(c2a); \draw[parr] (tc6)--(c2a);
  \draw[parr] (tc7)--(c2b); \draw[parr] (tc8)--(c2b);

  \node[font=\tiny,gray!62,rotate=90,align=center] at (5.15,2.65)
    {Hillis--Steele (intra)};

  \node[sbox={green!25},minimum width=1.0cm] (B1) at (0.93,1.8)
    {\tiny$u(B_1)$};
  \draw[parr] (c1a)--(B1) node[midway,left,font=\tiny]{$\oplus$};
  \draw[parr] (c1b)--(B1);

  \node[sbox={purple!25},minimum width=1.0cm] (B2) at (3.63,1.8)
    {\tiny$u(B_2)$};
  \draw[parr] (c2a)--(B2) node[midway,left,font=\tiny]{$\oplus$};
  \draw[parr] (c2b)--(B2);

  \node[mop] (mg) at (2.28,1.0) {$\oplus$};
  \draw[parr] (B1)--(mg);
  \draw[parr] (B2)--(mg);

  \node[font=\tiny,gray!62,rotate=90,align=center] at (5.15,0.5)
    {Blelloch (inter)};

  \node[sbox={blue!20},minimum width=1.1cm] (outc) at (2.28,0.3)
    {\tiny output $y$};
  \draw[->,thin] (mg)--(outc);

  \draw[<->,thick,blue!45] (5,1.6)--(5,3.72)
    
    node[midway, right=1pt, font=\tiny, blue!65!black, rotate=90, anchor=south]
      {$\lceil\log_2 B\rceil$};
  \draw[<->,thick,blue!45] (5,0.12)--(5,0.82)
    
    node[midway, right=1pt, font=\tiny, blue!65!black, rotate=90, anchor=south]
      {$\log\tfrac{n}{B}$};

  \node[font=\tiny,gray!65] at (2.28,-0.18)
    {Total: $L(n,B)=O(\log n)$\;\;(Eq.~\ref{eq:depth})};
\end{scope}

\end{tikzpicture}}
\vspace{-3mm}
\caption{%
\textbf{Integral-image analogy and ELSA computational flow.}
  \textbf{(a)}~Just as a prefix-sum table~\cite{crow1984} enables $O(1)$
  range queries by composing boundary values, the softmax state $(m,S,W)$
  supports exact blockwise composition via $\oplus$
  (Proposition~\ref{prop:block}).
  \textbf{(b)}~ME-SDPA~\cite{xformers2022} chains tokens sequentially via
  the online softmax recurrence~\cite{rabe2021}, incurring $O(n)$ depth
  that becomes prohibitive at long sequences.
  \textbf{(c)}~ELSA reduces depth to \textbf{$O(\log n)$}
  via a two-level prefix scan:
  \textbf{Hillis--Steele}~\cite{hillis1986} within
  blocks (all threads active, maximizing shared-memory bandwidth) and
  \textbf{Blelloch}~\cite{Blelloch1990} across blocks
  (work-optimal, minimizing global memory traffic)---while preserving
  \textbf{exact FP32 semantics}, $O(n)$ memory, and
  hardware agnosticism: $\oplus$ requires only scalar
  \texttt{max}/\texttt{exp} and vector \texttt{fma}, with
  \textbf{no Tensor Core dependency}.%
}

\vspace{-4mm}
\label{fig:overview}
\end{figure*}

\noindent\textbf{Efficient Memory and Architecture Design.}
Beyond approximation and hardware fusion, a parallel line of work reduces
memory through architectural redesign.
In distributed settings, DistFlashAttn~\cite{li2023distflashattn} overlaps
compute and communication to extend FA2 across devices.
In vision, representative designs include EfficientViT's cascaded group
attention~\cite{liu2023efficientvit}, ELFATT's sparse-blockify
approach~\cite{wu2025elfatt}, SHViT's single-head design~\cite{yun2024shvit},
L2ViT's information concentration~\cite{zheng2025l2vit}, and META's
cross-shaped window attention~\cite{zhang2025meta}.
While these designs achieve efficiency through structural modifications,
they require task-specific retraining and
\textbf{cannot serve as drop-in replacements} for
pretrained models.

Most closely related to ELSA are three lines of work.
Feng \emph{et al.}~\cite{feng2024} identify that the online softmax
recurrence admits parallel prefix reduction, but neither formalize it as a
monoid nor deliver a \textbf{deployable FP32 kernel}
with a quantitative error bound.
LeanAttention~\cite{sanovar2024lean} exploits associativity only for
single-query decode.
ME-SDPA~\cite{xformers2022} preserves exact FP32 semantics but processes
tokens sequentially, incurring $\mathcal{O}(n/T_k)$ depth and a
$3.5\times$ latency penalty at 16K tokens.
ELSA addresses all three gaps:
\textbf{(i)}~a \emph{triple-state monoid} $(m,S,W)$
with formally proven closure, associativity, and identity;
\textbf{(ii)}~\emph{all-queries} (prefill) attention;
and \textbf{(iii)}~a
\textbf{provable FP32 error bound} tied to scan depth
$L(n,B)$---making ELSA a general, training-free, numerically auditable
drop-in for vision and language inference.

%% file: sec/4-method.tex
\vspace{-3mm}
\section{Motivation and Preliminaries}
\label{sec:preliminary}
\vspace{-1mm}

Efficient attention requires addressing two intertwined challenges:
the $\mathcal{O}(n^2)$ memory cost of materializing the full score matrix,
and the sequential depth of memory-efficient implementations that avoid it.
We first characterize the sequential bottleneck and then introduce the
parallel prefix scan framework that ELSA exploits to overcome it.

\vspace{-1mm}
\subsection{Why Is Sequential Attention a Bottleneck?}

\vspace{-1mm}

\paragraph{Self-attention and Quadratic Memory.}
Given a sequence of $n$ tokens, let $q_i, k_j \in \mathbb{R}^d$ denote the
query and key vectors and $v_j \in \mathbb{R}^{d_v}$ the value vector.
The attention score $s_{ij} = \langle q_i, k_j \rangle / \sqrt{d}$ and
output are
\begin{equation}
y_i = \frac{\sum_{j} e^{s_{ij}} v_j}{\sum_{j} e^{s_{ij}}}.
\end{equation}
Naive materialization of the full $n{\times}n$ score matrix requires
$\mathcal{O}(n^2)$ memory, which becomes prohibitive at long sequence lengths.
To avoid this, one instead maintains a running state $(m_j, S_j, W_j)$,
where $m_j = \max_{u \le j} s_u$ is the running maximum logit,
$S_j = \sum_{u=1}^{j} e^{s_u - m_j}$ the normalized cumulative sum, and
$W_j = \sum_{u=1}^{j} e^{s_u - m_j} v_u$ the corresponding weighted sum,
reducing memory to $\mathcal{O}(n)$ without storing intermediate scores.

\paragraph{Online Softmax Recurrence and Its Sequential Depth.}
Rabe and Staats~\cite{rabe2021} showed this $\mathcal{O}(n)$-memory
computation can be performed in a single forward sweep via
\begin{equation}\label{eq:online-softmax}
\scalebox{0.88}{$\displaystyle
\begin{aligned}
m_j &= \max(m_{j-1},\, s_j), \\
S_j &= S_{j-1}\,e^{m_{j-1}-m_j} + e^{s_j-m_j}, \\
W_j &= W_{j-1}\,e^{m_{j-1}-m_j} + e^{s_j-m_j}\,v_j.
\end{aligned}
$}
\end{equation}
Heinsen~\cite{heinsen2024} later proposed a constant-cost-per-token variant
confirming the same memory property.
However, both formulations are \emph{inherently sequential}: each step
depends on the previous, so a GPU must synchronize $n$ times, leaving most
threads idle.
This $\mathcal{O}(n/T_k)$ sequential depth in memory-efficient kernels such
as ME-SDPA~\cite{xformers2022} is the core latency bottleneck we address.
Throughout, we use \emph{exact} to mean that our output equals standard
softmax attention in real arithmetic; in floating point, we derive a
provable FP32 relative error bound (Theorem~\ref{thm:fp32}).
\vspace{-1mm}
\subsection{Parallelization via Prefix Scan}
\vspace{-1mm}
\paragraph{Parallel Prefix Scanning.}
To overcome the sequential bottleneck, we turn to the theory of parallel
prefix scan~\cite{Blelloch1990}.
A prefix scan computes all prefix reductions over a monoid $(\oplus, e)$:
a set with an associative binary operator and an identity element.
Classic algorithms, Hillis--Steele~\cite{hillis1986},
Kogge--Stone~\cite{kogge1973}, and Blelloch's two-pass~\cite{Blelloch1990},
achieve \textbf{$\mathcal{O}(\log n)$ depth} with
\textbf{$\mathcal{O}(n)$ work}.
Modern GPU libraries (CUDA CUB, Thrust) generalize these primitives to
composite data types: as long as the tuple-level operation is associative
and has an identity, prefix scan accumulates vector- or tuple-valued states
in parallel \textbf{without Tensor Core dependencies}.
This raises a natural question: \emph{can the online softmax recurrence be
cast as such a monoid?} We answer affirmatively in Sec.~\ref{sec:methodology}.

\vspace{-2mm}
\paragraph{Integral-image View of Softmax Attention.}
A key obstacle is that each recurrence step depends on the \emph{global}
running maximum $m_j$, unknown until all previous tokens are processed.
The integral-image view resolves this: just as a 2D integral image lets any
rectangular region sum be retrieved in $\mathcal{O}(1)$ by combining corner
values, the attention state over any contiguous block can be recovered by
composing the states of its sub-blocks.
This composability means blocks can be computed \emph{independently in
parallel} and merged in a \textbf{logarithmic reduction tree}.

Concretely, let $s_j = \langle q, k_j \rangle / \sqrt{d}$ and
$m_j = \max_{u \le j} s_u$.
Define the unnormalized prefix quantities
\begin{equation}
\scalebox{0.88}{$\displaystyle
\tilde{S}_j = \sum_{u \le j} e^{s_u}, \qquad
\tilde{W}_j = \sum_{u \le j} e^{s_u}\, v_u,
$}
\end{equation}
and the normalized state
$(m_j, S_j, W_j) = (m_j,\, \tilde{S}_j e^{-m_j},\, \tilde{W}_j e^{-m_j})$.
Normalization by $e^{-m_j}$, analogous to the origin correction in integral
images, re-anchors accumulated sums to the local maximum, preventing
overflow while preserving exact values.
The maps $\textsf{un}(m,S,W) = (Se^m, We^m)$ and
$\textsf{renorm}(m,\tilde{S},\tilde{W}) = (m, \tilde{S}e^{-m}, \tilde{W}e^{-m})$
convert between representations; our merge operator $\oplus$ implements
blockwise composition via \textbf{``un-sum-renorm''},
the softmax analogue of summing corner values.
This yields a scan over $(m,S,W)$ in
\textbf{$\mathcal{O}(\log n)$ depth} with
\textbf{$\mathcal{O}(n)$ extra memory},
what we term \emph{linear-scan}.
The key question is whether this blockwise composability can be formalized
as a monoid, enabling standard parallel scan primitives.
We answer this in Sec.~\ref{sec:methodology}; the complete monoid proof
and I/O complexity analysis are deferred to the supplementary material.

\vspace{-2mm}
\section{Exact Linear-Scan Attention}
\label{sec:methodology}
\vspace{-1mm}

\begin{figure}[t]
\scalebox{0.8}{
\centering
\begin{tikzpicture}[
  >=stealth, semithick,
  sbox/.style={draw, rounded corners=2pt, fill=#1,
               minimum width=2.2cm, minimum height=0.45cm,
               font=\small, inner sep=3pt},
  sbox/.default={blue!10},
  arr/.style={->, thick, gray!70},
  obox/.style={draw, rounded corners=2pt, fill=orange!15,
               minimum width=1.6cm, minimum height=0.4cm,
               font=\small, inner sep=3pt},
  annot/.style={font=\scriptsize, gray!70, align=center},
]
\node[sbox={green!15}] (A) at (-1.8, 4.0)
  {$(m_a,\, S_a,\, W_a)$};
\node[font=\scriptsize, green!60!black] at (-1.8, 4.5) {Block $A$};
\node[sbox={purple!15}] (B) at (1.8, 4.0)
  {$(m_b,\, S_b,\, W_b)$};
\node[font=\scriptsize, purple!60!black] at (1.8, 4.5) {Block $B$};
\node[obox] (unA) at (-1.8, 2.9)
  {$S_a e^{m_a},\; W_a e^{m_a}$};
\node[obox] (unB) at (1.8, 2.9)
  {$S_b e^{m_b},\; W_b e^{m_b}$};
\draw[arr] (A)--(unA);
\draw[arr] (B)--(unB);
\node[annot] at (-3.3, 3.45) {\texttt{un}$(\cdot)$};
\node[annot] at (3.3, 3.45) {\texttt{un}$(\cdot)$};
\draw[dashed, gray!35] (-3.6, 2.45) -- (3.6, 2.45);
\node[font=\scriptsize, gray!50] at (3.2, 2.55) {add};
\node[sbox={yellow!20}] (add) at (0, 1.9) {%
  $\max(m_a,m_b),\;\tilde{S}_a{+}\tilde{S}_b,\;\tilde{W}_a{+}\tilde{W}_b$};
\draw[arr] (unA.south) .. controls (-2.8,2.3) and (-2.4,1.9) .. (add.west);
\draw[arr] (unB.south) .. controls (2.8,2.3) and (2.4,1.9) .. (add.east);
\node[annot] at (3.3, 1.45) {\texttt{renorm}$(\cdot)$};
\draw[arr] (add)--(0, 0.95);
\draw[dashed, gray!35] (-3.6, 1.45) -- (3.6, 1.45);
\node[font=\scriptsize, gray!50] at (-2.9, 1.55) {renorm};
\node[sbox={blue!18}] (out) at (0, 0.55)
  {$(m,\; S,\; W) = u(A \cup B)$};
\node[font=\scriptsize, blue!65!black] at (0, 0.05)
  {$= u_A \oplus u_B \quad$ (Eq.~\ref{eq:merge-compact})};
\draw[decorate, decoration={brace, amplitude=4pt},
      gray!50, thin]
  (3.65, 4.25) -- (3.65, 2.6)
  node[midway, right=4pt, font=\scriptsize, gray!65, align=left]
    {unnormalize\\(expose raw sums)};
\draw[decorate, decoration={brace, amplitude=4pt},
      gray!50, thin]
  (3.65, 2.3) -- (3.65, 1.6)
  node[midway, right=4pt, font=\scriptsize, gray!65, align=left]
    {element-wise\\add \& max};
\draw[decorate, decoration={brace, amplitude=4pt},
      gray!50, thin]
  (3.65, 1.3) -- (3.65, 0.35)
  node[midway, right=4pt, font=\scriptsize, gray!65, align=left]
    {renormalize\\by $e^{-m}$};
\end{tikzpicture}}
\vspace{-5mm}
\caption{\textbf{Monoid merge operator $\oplus$.}
Three steps compose two block states: \texttt{un} exposes raw sums,
element-wise $+$ and $\max$ aggregate them, and \texttt{renorm}
re-anchors to the new maximum.}
\label{fig:merge}
\vspace{-5mm}
\end{figure}
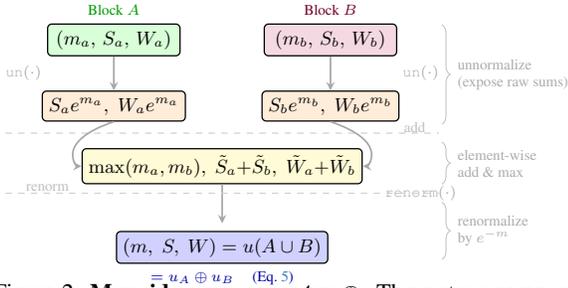

\paragraph{Overview.}
The integral-image view in Sec.~\ref{sec:preliminary} suggests that the
online softmax state $(m,S,W)$ can be composed blockwise (Figure~\ref{fig:overview}); Figure~\ref{fig:merge} details the
three-step merge operator $\oplus$ that implements this composition.
We begin by formalizing this observation.

\begin{proposition}[Block composition]\label{prop:block}
For any partition of indices into contiguous blocks $B_1, \dots, B_K$,
let $u(B_k)$ be the state $(m,S,W)$ computed on block $B_k$.
Then $u(B_1 \cup \cdots \cup B_K) = u(B_1) \oplus \cdots \oplus u(B_K)$.
\end{proposition}

\noindent This composability is the foundation of ELSA: because block
states can be computed \emph{independently in parallel} and merged in a
logarithmic reduction tree, it suffices to show that $\oplus$ forms a
\emph{monoid}---thereby admitting standard parallel scan primitives
(Sec.~\ref{sec:preliminary}, Parallel prefix scanning).
Unlike LeanAttention~\cite{sanovar2024lean}, which exploits a related
property only for single-query decode, our formulation covers all-queries
(prefill) attention and provides a formal FP32 error bound tied to scan
depth.
Similarly, while Feng \emph{et al.}~\cite{feng2024} identify this algebraic
structure theoretically, they do not provide a deployable high-throughput
FP32 kernel; our contribution lies precisely in this gap.

In this section, we proceed in three steps:
\textbf{(1)}~proving the monoid structure of $\oplus$,
\textbf{(2)}~deriving the FP32 error bound, and
\textbf{(3)}~designing the two-level GPU scan.

\subsection{Monoid Formulation}
Let $U = \bar{\mathbb{R}} \times \mathbb{R}_{\ge 0} \times \mathbb{R}^{d_v}$ with $\bar{\mathbb{R}} = \mathbb{R} \cup \{-\infty\}$, and
$\boldsymbol{e} = (-\infty, 0, \mathbf{0})$.
The merge operator $\oplus$ is designed to compose the running states of Eq.~\eqref{eq:online-softmax} blockwise. For $u_a = (m_a, S_a, W_a)$ and $u_b = (m_b, S_b, W_b)$, define
\vspace{-1mm}
\begin{equation}\label{eq:merge}
\scalebox{0.82}{$
u_a \oplus u_b =
\begin{cases}
  \bigl(m_a,\; S_a + S_b\,e^{m_b-m_a},\; W_a + W_b\,e^{m_b-m_a}\bigr), & m_a \ge m_b, \\[4pt]
  \bigl(m_b,\; S_b + S_a\,e^{m_a-m_b},\; W_b + W_a\,e^{m_a-m_b}\bigr), & \text{otherwise.}
\end{cases}
$}
\end{equation}

\vspace{-1mm}
\noindent\textbf{Closure.}
Since $e^{m_b - m_a} \in (0,1]$ when $m_a \ge m_b$, $S \in \mathbb{R}_{\ge 0}$ is preserved and $W \in \mathbb{R}^{d_v}$ is closed under affine combination; the symmetric case follows by construction.

\noindent\textbf{Associativity.} With $\textsf{un}(m,S,W) = (Se^m, We^m)$ and $\textsf{renorm}(m,\tilde{S},\tilde{W}) = (m, \tilde{S}e^{-m}, \tilde{W}e^{-m})$,
\begin{equation}\label{eq:merge-compact}
\scalebox{0.88}{$\displaystyle
u_a \oplus u_b = \textsf{renorm}\bigl(\max(m_a, m_b),\;
\textsf{un}(u_a) + \textsf{un}(u_b)\bigr).
$}
\end{equation}
\vspace{-1mm}
By associativity of vector addition and $\max$, $(u_a \oplus u_b) \oplus u_c = u_a \oplus (u_b \oplus u_c)$.

\noindent\textbf{Identity.} $(U, \oplus, \boldsymbol{e})$ forms a monoid with identity $\boldsymbol{e} = (-\infty, 0, \mathbf{0})$; a complete proof is given in Supp.~\ref{sec:Monoid}.

\subsection{Logical Scan and GPU Parallelization}

\textit{Logical level.}
Proposition~\ref{prop:block} guarantees that the state $(m,S,W)$ composes
associatively via $\oplus$ (Eq.~\eqref{eq:merge}), so the full-sequence
result equals the sequential prefix product $u_1 \oplus \cdots \oplus u_n$.
The per-query sequential formulation (detailed pseudocode and step-by-step explanation in Supp.~\ref{sec:algA4}) makes this explicit: it initializes
$(m_0, S_0, W_0) = (-\infty, 0, \mathbf{0})$, computes scores,
applies the conditional rescaling of Eq.~\eqref{eq:online-softmax}, and normalizes.
Although the loop is written sequentially, its correctness relies solely on
the monoid structure, not on sequential order, which is precisely what enables parallel execution.

\textit{Physical level.}
A naive single-level parallel scan would require all $n$ threads to
synchronize through global memory, incurring high communication overhead at
long sequences.
ME-SDPA~\cite{xformers2022} avoids this by processing tiles sequentially,
but at the cost of $\mathcal{O}(n/T_k)$ depth.
FA2/FA3~\cite{dao2023,shah2024flashattention} reduce depth via block tiling,
but advance \emph{sequentially} along the key dimension and depend on Tensor
Core fusion unavailable in FP32.
To achieve $\mathcal{O}(\log n)$ depth without these dependencies, we exploit the two-level structure of $\oplus$: since $\oplus$ is associative,
blocks can be reduced independently and their results merged in a
logarithmic tree.
Concretely, since $(U, \oplus, \boldsymbol{e})$ is a monoid, we map the computation to a two-level
prefix scan, choosing algorithms whose properties match each level's
constraints:
\begin{enumerate}
\item \textbf{Intra-block ($B = 128$):} Each thread block independently
reduces a chunk of 128 tokens via a \textbf{Hillis--Steele
scan}~\cite{hillis1986} in shared memory, contributing
\textbf{$\lceil \log_2 B \rceil$ merge levels}.
Because $\oplus$ is associative (Sec.~\ref{sec:methodology}, Monoid
Formulation), any ordering of pairwise merges within the block yields
the same result, making Hillis--Steele's tree-structured reduction
provably correct.
Hillis--Steele is preferred here because
\textbf{all $B$ threads remain active throughout
every level}, maximizing shared-memory bandwidth utilization;
at the small block size $B{=}128$, its $\mathcal{O}(B\log B)$ work
overhead over the $\mathcal{O}(B)$-optimal alternative is negligible
relative to the latency saved by keeping threads busy.
Block size $B{=}128$ balances occupancy and shared-memory usage across
GPU generations.

\item \textbf{Inter-block:} A \textbf{Blelloch two-pass
scan}~\cite{Blelloch1990} aggregates block-level states in global memory,
adding \textbf{$2\lceil \log_2(n/B) \rceil$ merge
levels}.
The identity element $\boldsymbol{e} = (-\infty, 0, \mathbf{0})$ of the
monoid enables Blelloch's down-sweep to initialize boundary states
correctly without special-casing empty prefixes.
Unlike Hillis--Steele, Blelloch's up-sweep/down-sweep is
\textbf{work-optimal} ($\mathcal{O}(n/B)$ total
operations), which is critical at the inter-block level where $n/B$ can
reach the hundreds and unnecessary global memory traffic would dominate
latency.
The up-sweep computes each block's total $\mathbf{u}_{\text{tot}}$; the
down-sweep distributes prefix results back and combines with each block's
internal state via $\oplus$, yielding the total depth $L(n,B)$
(Eq.~\eqref{eq:depth}).
\end{enumerate}

\subsection{Numerical Stability, Scan Depth, and I/O Scope}

\paragraph{Assumption 1 (underflow and exp-accuracy).}
Exponentials underflowing below the IEEE-754 single-precision subnormal range
are treated as zero; the library \texttt{exp} satisfies
$|\mathrm{fl}(\exp(x)) - \exp(x)| \le \kappa\,u\,\exp(x)$ for all $x \le 0$,
for a modest constant $\kappa$ independent of $n$.

\begin{lemma}[Bounded exponents]\label{lem:bounded-exp}
With running-max rescaling $m_j = \max(m_{j-1}, s_{ij})$, we have
$e^{s_{ij} - m_j} \le 1$ for all terms in each prefix.
\end{lemma}

\paragraph{Two-level Merge Depth.}
Let $B$ be the intra-block size ($B{=}128$). The two-level scan induces at most
\begin{equation}\label{eq:depth}
\scalebox{0.88}{$\displaystyle
L(n,B) = \lceil \log_2 B \rceil + 2\,\lceil \log_2(n/B) \rceil + 3
        \;\le\; 2\lceil \log_2 n \rceil + 3
$}
\end{equation}
merge levels along any data path.

\begin{theorem}[FP32 relative error bound]\label{thm:fp32}
Assume IEEE-754 single precision with unit roundoff $u$, finite logits, and
Assumption~1. Let $\hat{y}$ be the ELSA output and $y$ the exact output in
real arithmetic.
\begin{enumerate}[label=(\alph*), leftmargin=12pt]
\item \textbf{Single query.} By Higham's $(1{+}\delta)$ model
\cite{Higham2002Accuracy} and Lemma~\ref{lem:bounded-exp},
\begin{equation}
\scalebox{0.8}{$\displaystyle
\frac{\|\hat{y} - y\|_2}{\|y\|_2}
  \;\le\; u\,L(n,B) + \mathcal{O}(u^2)
  \;\le\; u\bigl(2\lceil \log_2 n \rceil + 3\bigr)
  + \mathcal{O}(u^2).
$}
\end{equation}

\item \textbf{All queries (blocked).} Each output row follows a path of
depth $\le L(n,B)$; the same bound applies per row.
\end{enumerate}
\end{theorem}

\begin{figure}[t]
\centering
\scalebox{0.9}{
\includegraphics[width=\columnwidth]{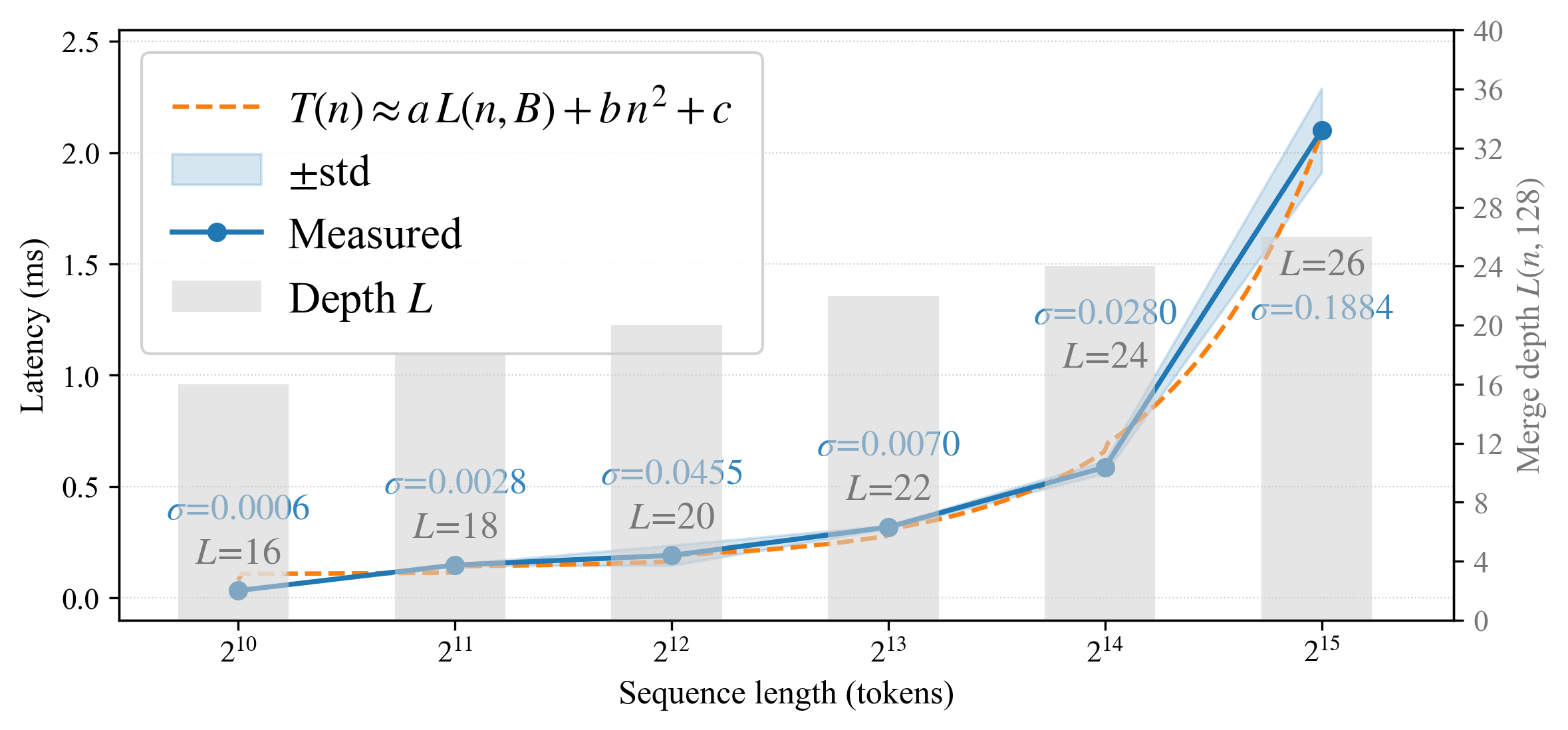}}
\vspace{-2mm}
\caption{\textbf{Per-head latency vs.\ sequence length.}
Measured latency fitted to $T(n)\approx a\,L(n,B)+b\,n^2+c$ (orange dashed);
vertical offsets track the logarithmic merge depth $L(n,128)$,
validating $\mathcal{O}(\log n)$ depth (Eq.~\ref{eq:depth}) and
$\mathcal{O}(n^2)$ work.}
\vspace{-4mm}
\label{fig:scaling}
\end{figure}

\paragraph{I/O scope.}
\emph{Per query:} ELSA streams $K, V$ once while updating $(m,S,W)$ and
writes one $y_i$, attaining the
\textbf{$\Omega(n)$ lower bound} for reading inputs
and writing the output.
\emph{All queries:} With tile sizes satisfying
$T_q d + T_k(d{+}d_v) \le cM_f$, the DRAM traffic is
\begin{equation}\label{eq:dram}
\scalebox{0.88}{$\displaystyle
\Theta(nd) + \Theta\!\left(\frac{n^2 d}{T_q}\right)
           + \Theta\!\left(\frac{n^2 d_v}{T_q}\right) + \Theta(nd_v)
$}
\end{equation}
and choosing $T_q \asymp \sqrt{M_f/(d{+}d_v)}$ yields
$\mathcal{O}(n^2(d{+}d_v)/\sqrt{M_f}) + \Theta(nd{+}nd_v)$,
\textbf{matching classical blocked I/O lower bounds}
up to constants; full derivations are in Supp.~\ref{sec:IO}.
\vspace{-3mm}
\paragraph{Empirical Scaling Validation.}
We measure per-head FP16 latency (median of 200 runs, single A100 GPU)
for $n\in\{2^{10},\dots,2^{15}\}$ tokens (Figure~\ref{fig:scaling}).
Latency fits closely to $T(n)\approx a\,L(n,B)+b\,n^2+c$
($a{=}0.0142$, $b{=}1.74{\times}10^{-9}$, $c{=}{-}0.149$, $B{=}128$),
confirming $\mathcal{O}(\log n)$ merge depth and $\mathcal{O}(n^2)$ overall
work, consistent with Theorem~1.

%% file: sec/5-experiment.tex
\definecolor{mygreen2}{RGB}{0,180,0}   
\definecolor{bestgray}{RGB}{220,220,220}   
\definecolor{secondgray}{RGB}{240,240,240} 

\vspace{-2mm}
\section{Experiments}
\label{sec:experiments}

\vspace{-2mm}
\paragraph{Implementation and Experimental Setup.}
We implement ELSA in Triton and CUDA C++ without HMMA/GMMA instructions,
making the kernels \emph{Tensor-Core independent} and deployable across GPU
generations.
Total arithmetic work remains $\mathcal{O}(n^2 d)$---identical to exact
attention---while the critical path is reduced to $\mathcal{O}(\log n)$,
an advantage that is most pronounced in FP32 where Tensor Core fusion is
not available.
Unless otherwise stated, benchmarks run on a single NVIDIA A100 (40GB,
CUDA~12.6, Triton~3.2.0, PyTorch~2.6), averaged over 200 runs; Jetson TX2
and host-device offloading results are reported separately.
We evaluate only \emph{exact}, drop-in replacements for standard multi-head
self-attention: the PyTorch Math kernel (Math), ME-SDPA in
xFormers~\cite{xformers2022}, FA2~\cite{dao2022} and
FA3~\cite{shah2024flashattention} where applicable, and ELSA (FP32 and
FP16).%
\footnote{\url{https://github.com/Dao-AILab/flash-attention}.
FA2/FA3 are evaluated in FP16/mixed precision as their FP32 kernels are
not competitive on our hardware.}
Approximate linear-time variants~\cite{Performer,Linformer,Zaheer2020} are
excluded as they modify the attention operator and require retraining.


\begin{figure}[h]
    \centering
    \includegraphics[width=0.48\textwidth]{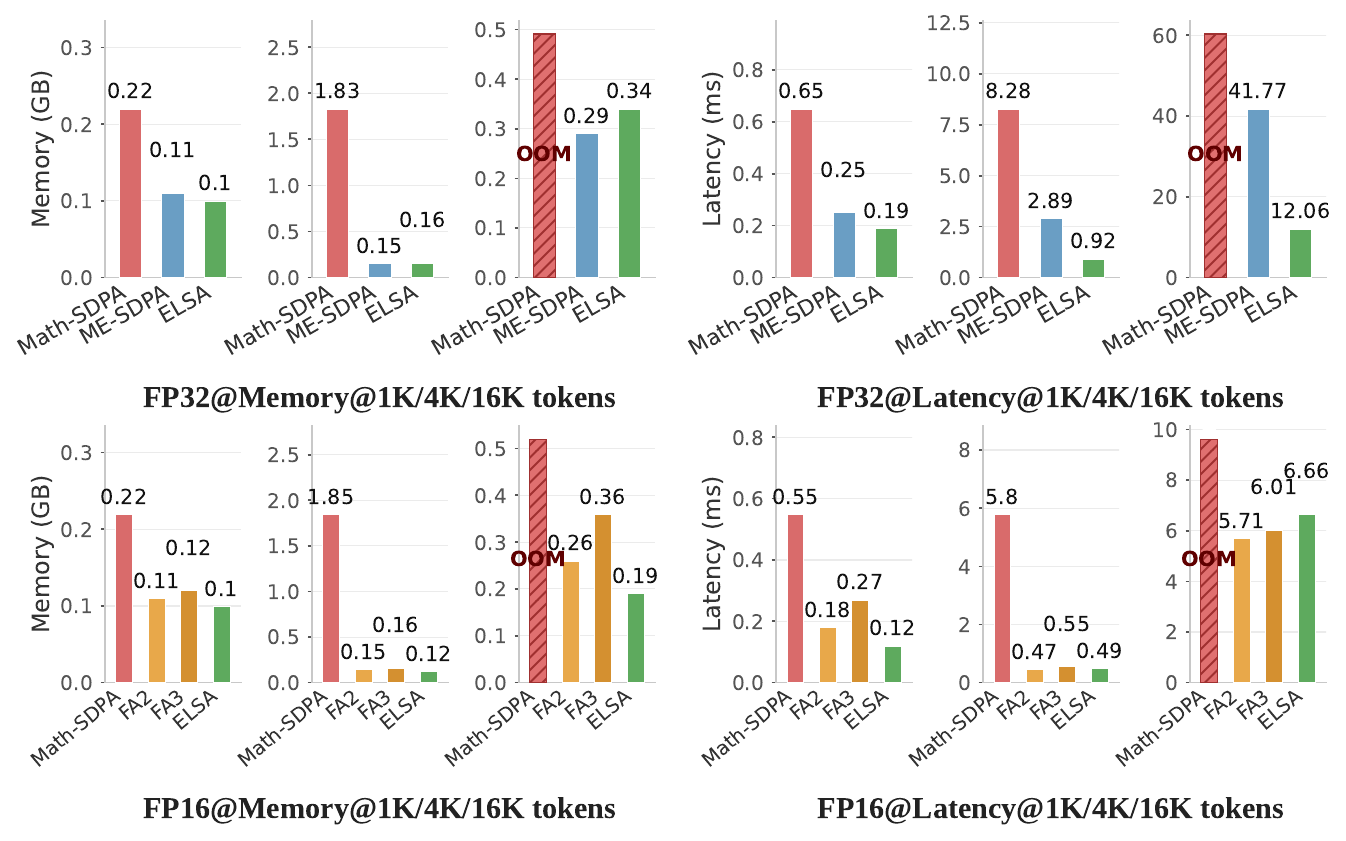}
    \vspace{-6mm}
\caption{\textbf{Latency and peak memory at 1K, 4K, and 16K tokens (FP16/FP32).}
In FP32, ELSA reduces latency over Math-SDPA and ME-SDPA while maintaining low memory.
In FP16, ELSA achieves the lowest memory among exact kernels and remains competitive
with FA2/FA3 at longer sequences; at 16K tokens, standard softmax attention fails
with OOM while ELSA remains stable.}
    \label{fig:table1}
    \vspace{-3mm}
\end{figure}

\colorlet{gradA}{bestgray!25}
\colorlet{gradB}{bestgray!50}
\colorlet{gradC}{bestgray!75}

\begin{table}[t]
\centering
\footnotesize
\scalebox{0.8}{%
\begin{tabular}{lcccccccc}
\toprule
& \multicolumn{4}{c}{\textbf{Speedup ($\times$)}} & \multicolumn{4}{c}{\textbf{VRAM Savings (\%)}} \\
\cmidrule(lr){2-5}\cmidrule(lr){6-9}
\textbf{Model} & 224 & 336 & 448 & 560
               & 224 & 336 & 448 & 560 \\
\midrule
ViT-B/16
  & 1.46
  & \cellcolor{gradA}1.63
  & \cellcolor{gradB}1.79
  & \cellcolor{gradC}\textbf{1.98}
  & 0.0
  & \cellcolor{gradA}2.3
  & \cellcolor{gradB}17.8
  & \cellcolor{gradC}\textbf{36.1} \\
ViT-B/32
  & 1.03
  & --
  & \cellcolor{gradB}1.42
  & --
  & 0.0
  & --
  & \cellcolor{gradB}0.0
  & -- \\
ViT-L/14
  & 1.54
  & \cellcolor{gradA}1.68
  & \cellcolor{gradB}1.87
  & \cellcolor{gradC}\textbf{2.12}
  & 0.0
  & \cellcolor{gradA}5.0
  & \cellcolor{gradB}20.8
  & \cellcolor{gradC}\textbf{39.6} \\
ViT-L/14-336
  & 1.56
  & \cellcolor{gradA}1.71
  & \cellcolor{gradB}1.90
  & \cellcolor{bestgray}\textbf{2.15}
  & 0.0
  & \cellcolor{gradA}5.0
  & \cellcolor{gradB}20.8
  & \cellcolor{bestgray}\textbf{39.6} \\
\bottomrule

\end{tabular}
}
\vspace{-2mm}
\caption{\textbf{ELSA vs.\ Math-SDPA on CLIP (\emph{attention-only proxy}, FP32):
  speedup and VRAM savings across resolutions.}
  Both gains scale with resolution; 
  Dashes indicate unevaluated configurations.
  For full image-encoder latency see Tab.~\ref{tab:clip}.}
\label{tab:clip-detailed}
\vspace{-4mm}
\end{table}
\vspace{-3mm}
\paragraph{Benchmarks and Scalability on Synthetic Sequences.}
We evaluate ELSA across a diverse set of tasks and models: ImageNet-1K
classification~\cite{Russakovsky2015} with ViT-B/16~\cite{dosovitskiy2021}
and Swin-T~\cite{liu2021}; zero-shot inference with CLIP
ViT-L/14~\cite{radford2021}; BERT sentiment on SST-2 and IMDB; and
hyperspectral classification (Pavia, Salinas, WHU) with
HSIMAE~\cite{wang2024hsimae}.
We benchmark single-GPU inference with random sequences of
$n=64$--$16{,}384$ tokens in both FP16 and FP32
(Figure~\ref{fig:table1}).
In FP16, at the isolated kernel level, ELSA remains close to FA2/FA3 in
latency at $n{=}4{,}096$, with the gap further narrowing beyond
$n{=}16{,}384$; this kernel-only gap reverses at the full-model level
(Table~\ref{tab:unified_vit_swin}).
FA2/FA3 are excluded from FP32 comparisons: their optimized paths rely on
HMMA/GMMA Tensor Core instructions for FP16/BF16, and their FP32 fallback
reverts to untuned SIMD execution, rendering the comparison uninformative.
More broadly, FA2/FA3 require Ampere/Hopper
Tensor Cores and are unavailable on older GPUs and edge
devices---deployment scenarios where ELSA operates as a hardware-agnostic,
FP32-exact drop-in replacement.
This resolution-scalability advantage is further evidenced in
Table~\ref{tab:clip-detailed}, which reports attention-module-only
resolution scaling (proxy scope) and should not be directly compared; extended results on 3D reconstruction
models (VGGT \cite{wang2025vggt} and FastVGGT \cite{shen2025fastvggt}) are provided in Supp.~\ref{sec:vggt_attention}, where ELSA achieves up to $2.34\times$
speedup over xFormers-FP32 while matching its memory footprint.
\vspace{-3mm}


  
  
  

  

\begin{table}[t]
\centering
\small
\setlength{\tabcolsep}{4pt}
\scalebox{0.72}{%
\begin{tabular}{llccc}
\toprule
\textbf{Model} & \textbf{Method} &
\textbf{Throughput (img/s)} &
\textbf{Peak Mem (MB)} &
\textbf{Mem vs Ref $\downarrow$} \\
\midrule
\multicolumn{5}{c}{\textbf{ViT Family (Global Attention, full-model, FP16, batch=8, strict scan only)}} \\
\midrule
\multirow{3}{*}{ViT-T}
 & Math-SDPA & 868.44 & 43.5 & -- \\
 & FA2       & 791.03 & 36.0 & $-17.2\%$ \\
 & ELSA      & \textbf{1309.24} & \textbf{33.8} & $\mathbf{-22.3\%}$ \\
\midrule
\multirow{3}{*}{ViT-S}
 & Math-SDPA & 897.03 & 96.4 & -- \\
 & FA2       & 859.73 & 79.8 & $-17.2\%$ \\
 & ELSA      & \textbf{1275.71} & \textbf{74.0} & $\mathbf{-23.2\%}$ \\
\midrule
\multirow{3}{*}{ViT-M}
 & Math-SDPA & 838.34 & 152.2 & -- \\
 & FA2       & 828.39 & 130.3 & $-14.4\%$ \\
 & ELSA      & \textbf{1204.31} & \textbf{113.3} & $\mathbf{-25.6\%}$ \\
\midrule
\multirow{3}{*}{ViT-B}
 & Math-SDPA & 820.91 & 282.3 & -- \\
 & FA2       & 787.39 & 249.6 & $-11.6\%$ \\
 & ELSA      & \textbf{1064.34} & \textbf{220.8} & $\mathbf{-21.8\%}$ \\
\midrule
\multicolumn{5}{c}{\textbf{Swin Family (Window Attention, full-model, FP16, batch=8, strict scan only)}} \\
\midrule
\multirow{3}{*}{Swin-T/W8}
 & WA        & 479.98 & \textbf{202.2} & -- \\
 & FA2       & 526.65 & 223.2 & $+10.4\%$ \\
 & W-ELSA    & \textbf{597.23} & 232.1 & $+14.8\%$ \\
\midrule
\multirow{3}{*}{Swin-T/W16}
 & WA        & 493.20 & \textbf{341.0} & -- \\
 & FA2       & 485.45 & 351.9 & $+3.2\%$ \\
 & W-ELSA    & \textbf{520.22} & 351.9 & $+3.2\%$ \\
\midrule
\multirow{3}{*}{Swin-S/W8}
 & WA        & 241.90 & \textbf{245.7} & -- \\
 & FA2       & 260.23 & 267.5 & $+8.9\%$ \\
 & W-ELSA    & \textbf{266.59} & 291.3 & $+18.6\%$ \\
\midrule
\multirow{3}{*}{Swin-S/W16}
 & WA        & 241.77 & 391.3 & -- \\
 & FA2       & 267.89 & 401.8 & $+2.7\%$ \\
 & W-ELSA    & \textbf{305.22} & \textbf{369.1} & $\mathbf{-5.7\%}$ \\
\bottomrule
\end{tabular}}
\caption{\textbf{ImageNet-1K full-model throughput and memory (FP16, batch${=}8$, strict scan only).}
  Full-model forward on a single A100-40GB at $224{\times}224$ with the same batch size
  (batch${=}8$) for ViT and Swin. All ELSA / W-ELSA rows use the corrected strict FP16
  two-scan path; no FP32 bridge and no legacy triton fast path are used.
  \texttt{ME-SDPA} is omitted: under FP16, \texttt{FA2} is the relevant flash baseline,
  while \texttt{Math-SDPA} (ViT) and \texttt{WA} (Swin) are retained only as
  memory-reference rows for the ``Mem vs Ref'' column. 
  Under this strict same-batch protocol, ELSA / W-ELSA is faster than FA2 on every
  reported ViT and Swin configuration, and also stays below the Math / WA memory
  reference on all ViT rows and on the Swin-S/W16 row.}
\label{tab:unified_vit_swin}
\vspace{-4mm}
\end{table}

\begin{figure}[t]
    \centering
    \includegraphics[width=1\linewidth]{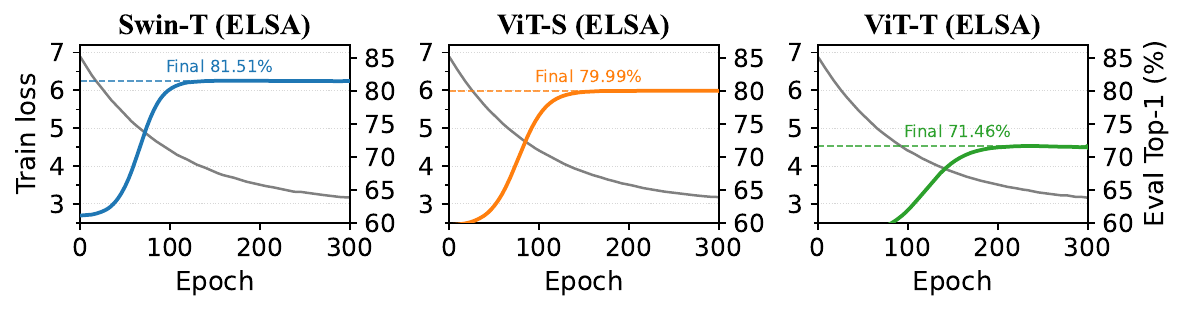}
    \vspace{-5mm}
    \caption{\textbf{Training convergence on ImageNet-1K.}
      Training loss (gray) and top-1 accuracy (colored) for classification models
      with ELSA. Dashed lines indicate peak accuracy.
      All models converge smoothly, confirming that ELSA does not disturb optimization.}
    \label{fig:training-converge}
    \vspace{-5mm}
\end{figure}

\paragraph{Training and Inference on ImageNet-1K Classification.}
Table~\ref{tab:unified_vit_swin} reports full-model FP16 forward at the
same batch size (batch${=}8$) for ViT and Swin~\cite{liu2021} using the
corrected strict FP16 scan path (no FP32 bridge, no legacy triton).
We drop \texttt{ME-SDPA} since \texttt{FA2} is the relevant FP16 flash
baseline; \texttt{Math-SDPA}/\texttt{WA} are kept only as memory
references. Under this strict same-batch protocol, ELSA\,/\,W-ELSA is
faster than FA2 on every ViT and Swin row (${\sim}29$--$65\%$ on ViT,
${\sim}2$--$14\%$ on Swin), and also stays below the Math/WA memory
reference on all ViT rows and on Swin-S/W16; the three remaining Swin
rows incur moderate memory overhead (${+}3.2\%$ to ${+}18.6\%$) from
short-window scan workspace, reported explicitly.
Beyond inference, ELSA deploys seamlessly as a
drop-in replacement during training with no modifications to the
optimization pipeline.
As shown in Figure~\ref{fig:training-converge}, Swin-T, ViT-S, and ViT-T
trained entirely with ELSA exhibit steady loss decay and monotonically
improving top-1 accuracy, with final values within 0.1\% of the corresponding Math/FA2 baseline,
confirming no optimization penalty.
\vspace{-1mm}

\begin{table*}[t]
\centering
\resizebox{\textwidth}{!}{
\begin{tabular}{l|lcccccc}
\hline
                                 &                                   & \multicolumn{2}{c}{\textbf{Pavia}}                                               & \multicolumn{2}{c}{\textbf{Salinas}}                                              & \multicolumn{2}{c}{\textbf{WHU}}                                                 \\
\multirow{-2}{*}{\textbf{Model}} & \multirow{-2}{*}{\textbf{Method}} & \textbf{Throughput}                       & \textbf{Peak Mem. (GB)}              & \textbf{Throughput}                        & \textbf{Peak Mem. (GB)}              & \textbf{Throughput}                       & \textbf{Peak Mem. (GB)}              \\ \hline

\multirow{2}{*}{\textbf{HSIMAE-B}}
 & ME-SDPA                                & {$10499.20 \pm 14.19$ (-)} & {$0.11 \pm 0.00$} & 
                                      {$10427.11 \pm 18.20$ (-)} & {$0.10 \pm 0.00$} & 
                                      {$10631.04 \pm 13.62$ (-)} & {$0.11 \pm 0.00$} \\


 & \cellcolor{gray!15}\textbf{ELSA (Ours)}                     & 
   \cellcolor{gray!15}\textbf{$14401.45 \pm 53.37$ \cellcolor{gray!15}\textcolor{ForestGreen}{(+37.1\%)}} & \cellcolor{gray!15}\textbf{$0.11 \pm 0.00$} &
   \cellcolor{gray!15}\textbf{$14423.96 \pm 106.10$ \textcolor{ForestGreen}{(+38.3\%)}} & \textbf{$0.10 \pm 0.00$} &
   \cellcolor{gray!15}\textbf{$14928.86 \pm 15.66$ \textcolor{ForestGreen}{(+40.4\%)}} & \cellcolor{gray!15}\textbf{$0.11 \pm 0.00$} \\ \hline

\multirow{2}{*}{\textbf{HSIMAE-L}}
 & ME-SDPA                                & {$6471.18 \pm 97.38$ (-)}            & {$0.23 \pm 0.05$} &
                                      {$6489.98 \pm 20.87$ (-)}            & {$0.23 \pm 0.00$} & 
                                      {$6588.37 \pm 5.48$ (-)}             & {$0.23 \pm 0.00$} \\


 & \cellcolor{gray!15}\textbf{ELSA (Ours)}                     & 
  \cellcolor{gray!15} \textbf{$10478.84 \pm 91.26$ \textcolor{ForestGreen}{(+62.0\%)}} & \cellcolor{gray!15}\textbf{$0.25 \pm 0.00$} &
  \cellcolor{gray!15} \textbf{$10395.81 \pm 27.00$ \textcolor{ForestGreen}{(+60.2\%)}} & \cellcolor{gray!15}\textbf{$0.25 \pm 0.00$} &
   \cellcolor{gray!15} \textbf{$10657.24 \pm 7.37$ \textcolor{ForestGreen}{(+61.7\%)}}  & \cellcolor{gray!15}\textbf{$0.24 \pm 0.00$} \\ \hline

\end{tabular}}
\vspace{-3mm}
\caption{\textbf{Efficiency evaluation on hyperspectral classification task (FP32).}
  Throughput (imgs/s) and peak memory for HSIMAE-B/L~\cite{wang2024hsimae}
  on Pavia, Salinas, and WHU~\cite{zhong2021spectral}.
  ELSA consistently outperforms ME-SDPA in throughput with negligible
  memory overhead.}

\vspace{-5mm}
\label{tab:HSIMAE}
\end{table*}

\begin{table}[t]
\centering
\renewcommand{\arraystretch}{1.28}   
\setlength{\tabcolsep}{4pt}
\resizebox{0.85\linewidth}{!}{
\begin{tabular}{l| l c c}
\hline
\textbf{Task} & \textbf{Method} & \textbf{Latency (ms)} & \textbf{Memory (GB)}\\
\hline

\multirow{2}{*}{DPT COCO (FP32)}
 & ME-SDPA
   & 295.7 
   & 5.19 \\

 & \cellcolor{gray!15}\textbf{ELSA (Ours)} 
   & \cellcolor{gray!15}\textbf{252.7 {\color{ForestGreen}{(+14.6\%)}}}
   & \cellcolor{gray!15}5.49 \\
\hline

\multirow{2}{*}{BERT (FP16)}
 & FA2 
   & 224.1 
   & 3.67 \\

 & \cellcolor{gray!15}\textbf{ELSA (Ours)} 
   & \cellcolor{gray!15}\textbf{96.8 {\color{ForestGreen}{(+56.8\%)}}}
   & \cellcolor{gray!15}\textbf{0.37} \\
\hline

\multirow{2}{*}{Swin COCO (FP16)}
 & FA2 
   & 46.0 
   & 0.95 \\

 & \cellcolor{gray!15}\textbf{ELSA (Ours)} 
   & \cellcolor{gray!15}\textbf{40.1 {\color{ForestGreen}{(+12.8\%)}}}
   & \cellcolor{gray!15}\textbf{0.87} \\
\hline

\end{tabular}}
\vspace{-3mm}
\caption{\textbf{Downstream tasks inference (FP16 and FP32).}
Latency and peak memory for DPT depth estimation, BERT~\cite{devlin2019bert}
sentiment (SST-2/IMDB~\cite{socher2013,maas2011}), and Swin-based COCO
detection~\cite{lin2014}.
ELSA reduces latency without retraining on common devices.}
\vspace{-3mm}
\label{tab:downstream1}
\end{table}
\vspace{-4mm}
\paragraph{Downstream Tasks and LLM Inference.}
Table~\ref{tab:HSIMAE} reports FP32 throughput and peak memory for
HSIMAE-B/L on three hyperspectral datasets; ELSA consistently exceeds
ME-SDPA in throughput while remaining within the same sub-gigabyte memory
regime.
For NLP workloads, Table~\ref{tab:bert} (left) reports fixed-shape bucketed
BERT attention under strict FP32, where ELSA achieves $1.97\times$ (SST-2)
and $2.27\times$ (IMDB) speedup over ME-SDPA; Table~\ref{tab:bert} (right)
reports synthetic padding-free variable-length throughput across precision
variants.
To assess generality beyond vision, we evaluate ELSA on
LLaMA-13B~\cite{touvron2023llama} under host-device offloading
(Supp.~\ref{sec:CUDA}, Table~\ref{tab:offload_filled}): at long contexts
($\ge$32K tokens), ELSA delivers
\textcolor{cvprblue}{\textbf{17.8--20.2\% throughput gains}} over SDPA
with no retraining or weight modification; at shorter contexts ($\le$8K),
scan overhead outweighs the copy-overlap benefit.

\vspace{-4mm}

\begin{table}[t]
  \centering
  \setlength{\tabcolsep}{4pt}
  \resizebox{0.44\textwidth}{!}{
    \begin{tabular}{c|cccc|cccc}
    \hline
    \multirow{2}{*}{\textbf{Tokens}} &
    \multicolumn{4}{c|}{\textbf{FP16}} &
    \multicolumn{4}{c}{\textbf{FP32}} \\
    & \textbf{Math} & \cellcolor{gray!20}\textbf{ELSA} & \textbf{Ratio} & \textbf{Alloc (MB)}
    & \textbf{Math} & \cellcolor{gray!20}\textbf{ELSA} & \textbf{Ratio} & \textbf{Alloc (MB)} \\
    \hline

64  & 1.06 & \cellcolor{gray!20}{\textbf{0.69} \textcolor{ForestGreen}{\textbf{(+34.9\%)}}} & \textcolor{ForestGreen}{1.54$\times$} & 0.04
     & 1.06 & \cellcolor{gray!20}{\textbf{0.69} \textcolor{ForestGreen}{\textbf{(+34.9\%)}}} & \textcolor{ForestGreen}{1.54$\times$} & 0.08 \\

100 & 2.64 & \cellcolor{gray!20}{\textbf{1.71} \textcolor{ForestGreen}{\textbf{(+35.2\%)}}} & \textcolor{ForestGreen}{1.54$\times$} & 0.06
     & 2.64 & \cellcolor{gray!20}{\textbf{1.71} \textcolor{ForestGreen}{\textbf{(+35.2\%)}}} & \textcolor{ForestGreen}{1.54$\times$} & 0.12 \\

144 & 5.47 & \cellcolor{gray!20}{\textbf{3.53} \textcolor{ForestGreen}{\textbf{(+35.5\%)}}} & \textcolor{ForestGreen}{1.55$\times$} & 0.09
     & 5.49 & \cellcolor{gray!20}{\textbf{3.54} \textcolor{ForestGreen}{\textbf{(+35.6\%)}}} & \textcolor{ForestGreen}{1.55$\times$} & 0.18 \\

196 & 10.10 & \cellcolor{gray!20}{\textbf{6.47} \textcolor{ForestGreen}{\textbf{(+35.9\%)}}} & \textcolor{ForestGreen}{1.56$\times$} & 0.12
     & 10.12 & \cellcolor{gray!20}{\textbf{6.48} \textcolor{ForestGreen}{\textbf{(+36.0\%)}}} & \textcolor{ForestGreen}{1.56$\times$} & 0.24 \\

256 & 16.74 & \cellcolor{gray!20}{\textbf{10.52} \textcolor{ForestGreen}{\textbf{(+37.2\%)}}} & \textcolor{ForestGreen}{1.59$\times$} & 0.16
     & 16.77 & \cellcolor{gray!20}{\textbf{10.54} \textcolor{ForestGreen}{\textbf{(+37.1\%)}}} & \textcolor{ForestGreen}{1.59$\times$} & 0.31 \\

324 & 27.22 & \cellcolor{gray!20}{\textbf{17.24} \textcolor{ForestGreen}{\textbf{(+36.6\%)}}} & \textcolor{ForestGreen}{1.58$\times$} & 0.20
     & 27.29 & \cellcolor{gray!20}{\textbf{17.28} \textcolor{ForestGreen}{\textbf{(+36.7\%)}}} & \textcolor{ForestGreen}{1.58$\times$} & 0.40 \\

400 & 41.23 & \cellcolor{gray!20}{\textbf{25.95} \textcolor{ForestGreen}{\textbf{(+37.0\%)}}} & \textcolor{ForestGreen}{1.59$\times$} & 0.24
     & 41.34 & \cellcolor{gray!20}{\textbf{26.04} \textcolor{ForestGreen}{\textbf{(+36.9\%)}}} & \textcolor{ForestGreen}{1.59$\times$} & 0.49 \\

484 & 60.01 & \cellcolor{gray!20}{\textbf{37.63} \textcolor{ForestGreen}{\textbf{(+37.3\%)}}} & \textcolor{ForestGreen}{1.59$\times$} & 0.30
     & 60.16 & \cellcolor{gray!20}{\textbf{37.74} \textcolor{ForestGreen}{\textbf{(+37.3\%)}}} & \textcolor{ForestGreen}{1.59$\times$} & 0.59 \\

576 & 84.54 & \cellcolor{gray!20}{\textbf{52.78} \textcolor{ForestGreen}{\textbf{(+37.6\%)}}} & \textcolor{ForestGreen}{1.60$\times$} & 0.35
     & 84.76 & \cellcolor{gray!20}{\textbf{52.96} \textcolor{ForestGreen}{\textbf{(+37.5\%)}}} & \textcolor{ForestGreen}{1.60$\times$} & 0.70 \\

676 & 116.85 & \cellcolor{gray!20}{\textbf{73.03} \textcolor{ForestGreen}{\textbf{(+37.5\%)}}} & \textcolor{ForestGreen}{1.60$\times$} & 0.41
     & 117.18 & \cellcolor{gray!20}{\textbf{73.33} \textcolor{ForestGreen}{\textbf{(+37.4\%)}}} & \textcolor{ForestGreen}{1.60$\times$} & 0.83 \\

784 & 157.45 & \cellcolor{gray!20}{\textbf{98.47} \textcolor{ForestGreen}{\textbf{(+37.4\%)}}} & \textcolor{ForestGreen}{1.60$\times$} & 0.48
     & 157.89 & \cellcolor{gray!20}{\textbf{98.86} \textcolor{ForestGreen}{\textbf{(+37.4\%)}}} & \textcolor{ForestGreen}{1.60$\times$} & 0.96 \\

900 & 207.55 & \cellcolor{gray!20}{\textbf{129.74} \textcolor{ForestGreen}{\textbf{(+37.5\%)}}} & \textcolor{ForestGreen}{1.60$\times$} & 0.55
     & 208.91 & \cellcolor{gray!20}{\textbf{131.14} \textcolor{ForestGreen}{\textbf{(+37.2\%)}}} & \textcolor{ForestGreen}{1.59$\times$} & 1.10 \\
    \hline
  \end{tabular}
}
  \vspace{-2mm}
\caption{\textbf{Embedded-system benchmark on Jetson TX2 (FP16 and FP32).}
  ELSA reduces latency by ${\sim}1.5\times$ over the Math baseline
  across all sequence lengths with no Tensor Core dependency.
  FP16 and FP32 latencies are nearly identical, confirming that TX2
  executes both precisions via SIMD rather than Tensor Cores.}
\vspace{-3mm}
\label{tab:Tx2-FP32}
\end{table}

\begin{table}[t]
\centering
\begin{minipage}[t]{0.48\linewidth}
  \centering
  \scalebox{0.53}{
  \hspace{-1mm}
  \begin{tabular}{lcccc}
  \toprule
  \textbf{Dataset}
    & \textbf{ME-SDPA}
    & \textbf{ELSA}
    & \textbf{Speedup}
    & \textbf{Lat.} \\
  \midrule
  SST-2
    & 1.308
    & \cellcolor{gray!15}\textbf{0.664}
    & \cellcolor{gray!15}\textcolor{ForestGreen}{\textbf{1.97$\times$}}
    & \cellcolor{gray!15}\textcolor{ForestGreen}{\textbf{49.3\%}} \\
  IMDB
    & 17.693
    & \cellcolor{gray!25}\textbf{7.794}
    & \cellcolor{gray!25}\textcolor{ForestGreen}{\textbf{2.27$\times$}}
    & \cellcolor{gray!25}\textcolor{ForestGreen}{\textbf{56.0\%}} \\
  \bottomrule
  \end{tabular}}
\end{minipage}
\hfill
\begin{minipage}[t]{0.49\linewidth}
  \centering
  \scalebox{0.53}{
  \begin{tabular}{lcc}
  \toprule
  \textbf{Method}
    & \textbf{M\,tok/s}
    & \textbf{vs baseline} \\
  \midrule
  SDPA-FP32-mem & 0.960 & -- \\
  \midrule
  \rowcolor{gray!10}ELSA-Strict (FP32) & \textbf{2.004} & \textcolor{ForestGreen}{\textbf{$+$108.7\%}} \\
  \rowcolor{gray!18}ELSA-Turbo (TF32)  & \textbf{2.643} & \textcolor{ForestGreen}{\textbf{$+$175.3\%}} \\
  \rowcolor{gray!28}ELSA-FP16  (FP16)  & \textbf{5.030} & \textcolor{ForestGreen}{\textbf{$+$424.0\%}} \\
  \bottomrule
  \end{tabular}}
  
\end{minipage}
\vspace{1mm}
\caption{\textbf{NLP benchmarks (strict FP32, TF32-Turbo, and FP16).}
  \textit{Left}: BERT fixed-shape bucketed attention vs.\ ME-SDPA;
  gains on IMDB reflect longer sequences.
  \textit{Right}: padding-free variable-length throughput across precision
  variants; shading reflects gain magnitude over SDPA-FP32-mem baseline.}
\label{tab:bert}
\label{tab:varlen_throughput}
\vspace{-6mm}
\end{table}

\paragraph{Evaluation on Embedded Systems.}
Table~\ref{tab:Tx2-FP32} evaluates ELSA on the resource-constrained Jetson
TX2, demonstrating its hardware-agnostic nature: since $\oplus$ requires
only scalar \texttt{max}/\texttt{exp} and vector \texttt{fma}---with no
Tensor Core or architecture-specific primitives---ELSA
deploys without modification on compute capability~6.2 where FA2 provides
no compatible FP32 kernel and FA3 is entirely unavailable.
Across 64--900 tokens (batch~1, $d{=}64$, FP16/FP32),
ELSA consistently reduces latency by
${\approx}35$--$38\%$ over the PyTorch Math kernel with identical memory,
confirming that the scan-based efficiency advantage transfers to edge
deployments. This makes ELSA particularly suited for latency-sensitive on-device inference scenarios, such as robotics and autonomous systems, where
Tensor Core--dependent kernels are unavailable.
\vspace{-4mm}

\paragraph{Fixed-Shape Bucketed BERT.}
On BERT attention workloads with bucketed fixed-shape padding,
ELSA consistently outperforms both ME-SDPA and Math-SDPA in strict FP32
(Table~\ref{tab:bert}).
On SST-2, ELSA achieves \textcolor{cvprblue}{\textbf{1.97$\times$}} speedup
over ME-SDPA;
on IMDB, \textcolor{cvprblue}{\textbf{2.27$\times$}} speedup.
The larger gain on IMDB reflects its longer average sequence length,
consistent with ELSA's scan efficiency scaling with $n$.
\vspace{-4mm}

\paragraph{Synthetic Padding-Free Variable-Length.}
Under a padding-free synthetic workload with variable-length packing
(Table~\ref{tab:varlen_throughput}),
ELSA-Strict (FP32) achieves \textcolor{cvprblue}{\textbf{2.00M\,tok/s}}.
ELSA-Turbo (TF32) further reaches $2.64$M\,tok/s ($+175.3\%$), and
ELSA-FP16 peaks at $5.03$M\,tok/s ($+424.0\%$), demonstrating that
the scan formulation's gains compound across precision modes.
\vspace{-4mm}
\paragraph{CLIP Image-Encoder Evaluation.}
Table~\ref{tab:clip} reports CLIP image-encoder latency (strict FP32 and
TF32-Turbo) across four ViT sizes. ELSA consistently reduces latency
over ME-SDPA (see Supp.~\ref{sec:Variants}), reaching $1.060\times$ at
ViT-L/14@336px (strict FP32) and $1.062\times$ via ELSA-Turbo;
per-variant details are in Supp.~\ref{sec:Variants}.
Since FA2~\cite{dao2023} and FA3~\cite{shah2024flashattention}
provide no compatible FP32 path for CLIP, ELSA is
the only hardware-agnostic exact-attention kernel that reduces parallel depth to $O(\log n)$ at full precision.

\begin{table}[t]
\centering
\scalebox{0.62}{
\begin{tabular}{l l c c}
\hline
\textbf{Model} & \textbf{Method} & \textbf{Latency (ms)} & \textbf{Speedup vs ME-SDPA}\\
\hline

\multirow{3}{*}{ViT-B/32}
 & ME-SDPA (baseline) & 2.398 & 1.00$\times$\\
 & \cellcolor{gray!15}ELSA-Strict (Ours) & \cellcolor{gray!15}\textbf{2.159}\;\;\textcolor{ForestGreen}{($-$10.0\%)} & \cellcolor{gray!15}\textbf{1.111$\times$}\\
 & \cellcolor{gray!25}ELSA-Turbo (Ours)  & \cellcolor{gray!25}\textbf{2.150}\;\;\textcolor{ForestGreen}{($-$10.3\%)} & \cellcolor{gray!25}\textbf{1.115$\times$}\\
\hline

\multirow{3}{*}{ViT-B/16}
 & ME-SDPA (baseline) & 4.568 & 1.00$\times$\\
 & \cellcolor{gray!15}ELSA-Strict (Ours) & \cellcolor{gray!15}\textbf{4.365}\;\;\textcolor{ForestGreen}{($-$4.4\%)}  & \cellcolor{gray!15}\textbf{1.046$\times$}\\
 & \cellcolor{gray!25}ELSA-Turbo (Ours)  & \cellcolor{gray!25}\textbf{4.329}\;\;\textcolor{ForestGreen}{($-$5.2\%)}  & \cellcolor{gray!25}\textbf{1.055$\times$}\\
\hline

\multirow{3}{*}{ViT-L/14}
 & ME-SDPA (baseline) & 15.419 & 1.00$\times$\\
 & \cellcolor{gray!15}ELSA-Strict (Ours) & \cellcolor{gray!15}\textbf{14.914}\;\;\textcolor{ForestGreen}{($-$3.3\%)} & \cellcolor{gray!15}\textbf{1.034$\times$}\\
 & \cellcolor{gray!25}ELSA-Turbo (Ours)  & \cellcolor{gray!25}\textbf{14.842}\;\;\textcolor{ForestGreen}{($-$3.7\%)} & \cellcolor{gray!25}\textbf{1.039$\times$}\\
\hline

\multirow{3}{*}{ViT-L/14@336px}
 & ME-SDPA (baseline) & 29.665 & 1.00$\times$\\
 & \cellcolor{gray!15}ELSA-Strict (Ours) & \cellcolor{gray!15}\textbf{27.989}\;\;\textcolor{ForestGreen}{($-$5.7\%)} & \cellcolor{gray!15}\textbf{1.060$\times$}\\
 & \cellcolor{gray!25}ELSA-Turbo (Ours)  & \cellcolor{gray!25}\textbf{27.931}\;\;\textcolor{ForestGreen}{($-$5.8\%)} & \cellcolor{gray!25}\textbf{1.062$\times$}\\
\hline

\end{tabular}}
\vspace{-2mm}
\caption{\textbf{CLIP image-encoder inference (image\_encoder\_only scope, strict FP32 and TF32-Turbo).}
  Latency measured on the full image encoder pipeline; speedup relative to
  ME-SDPA baseline. Memory is not reported in this table because it was
  not measured under this end-to-end image-encoder scope. Numbers are
  \emph{not} comparable to attention-only proxy results
  (cf.\ Tab.~\ref{tab:clip-detailed}).}
\vspace{-4mm}
\label{tab:clip}
\end{table}


%% file: sec/6-conclusion.tex
\section{Discussions}
\paragraph{Limitations.}
ELSA reduces memory and I/O to $O(n)$ but preserves quadratic arithmetic
complexity, and is most effective when memory rather than compute is the
bottleneck.
At the isolated kernel level on short sequences, scan merge overhead can
leave ELSA behind FA2/FA3, though this reverses at the full-model level
(Table~\ref{tab:unified_vit_swin}); under offloading, gains of
17.8--20.2\% over SDPA emerge only at $\ge$32K tokens.
The kernel is validated primarily on Ampere GPUs, and training dynamics
under multi-GPU scaling remain open.
\vspace{-3mm}
\paragraph{Future Directions.}
Promising avenues include incorporating sparsity or low-rank structures
to reduce FLOPs, extending to distributed multi-GPU training and
parallelization, broadening hardware coverage beyond Ampere.

\section{Conclusion}
\label{sec:conclusion}
We reframe online softmax attention as a prefix scan over an associative
monoid $(m,S,W)$ and present ELSA, a deployable \underline{strict FP32 exact
attention} kernel with a provable
\textbf{$\mathcal{O}(u\log n)$ relative error bound}.
ELSA retains $O(n^2)$ arithmetic cost while reducing
\textbf{parallel depth to $O(\log n)$}, uses $O(n)$
extra memory, and streams each query once, attaining the per-query
$\Omega(n)$ I/O lower bound.
Requiring no Tensor Core primitives, it runs identically on A100s and
Jetson TX2---a \underline{hardware-agnostic drop-in replacement} for
pretrained models that need neither retraining nor architectural
modification.
Within the strict FP32 exact-attention setting, where FA2/FA3 provide no
compatible path, ELSA consistently outperforms sequential exact baselines,
delivering \textcolor{cvprblue}{\textbf{$1.3$--$3.5\times$ speedups}} over
ME-SDPA across long sequences, and also surpasses FA2 on every ViT/Swin
row in the FP16 full-model same-batch regime
(Table~\ref{tab:unified_vit_swin}), making it a precise, memory-light
solution for high-precision vision and language inference.
Additional FP16/TF32 ablations and system-level profiling (offloading, CUDA C++ portability) are provided in the supplementary so as not to dilute the strict FP32 headline result.

%% file: sec/7-suppl.tex
\setcounter{page}{1}
\maketitlesupplementary

\definecolor{mygreen2}{RGB}{0,180,0}
\definecolor{bestgray}{RGB}{220,220,220}
\definecolor{secondgray}{RGB}{240,240,240}

\section*{Supplementary Overview}

This supplementary provides theoretical foundations, additional experimental
results, and implementation details for ELSA, organized as follows.

\bigskip
\renewcommand{\arraystretch}{1.4}
\noindent
\begin{tabular}{@{}l@{\hspace{.6em}}p{0.76\linewidth}r@{}}

\multicolumn{3}{@{}l}{\textbf{Theory and Analysis}} \\
\S\,\ref*{sec:Monoid}
  & \hyperref[sec:Monoid]{Monoid Structure and Scan Correctness}\dotfill
  & \pageref{sec:Monoid} \\
\S\,\ref*{sec:IO}
  & \hyperref[sec:IO]{I/O Complexity and Algorithmic Bounds}\dotfill
  & \pageref{sec:IO} \\
\S\,\ref*{sec:VecEq}
  & \hyperref[sec:VecEq]{Numerical Equivalence Verification}\dotfill
  & \pageref{sec:VecEq} \\[6pt]

\multicolumn{3}{@{}l}{\textbf{Additional Experiments}} \\
\S\,\ref*{sec:ablation}
  & \hyperref[sec:ablation]{Ablation Studies}\dotfill
  & \pageref{sec:ablation} \\
\S\,\ref*{sec:varlen}
  & \hyperref[sec:varlen]{Variable-Length Sequence Benchmark}\dotfill
  & \pageref{sec:varlen} \\
\S\,\ref*{sec:window_att}
  & \hyperref[sec:window_att]{Window Attention and Offloading Analysis}\dotfill
  & \pageref{sec:window_att} \\
\S\,\ref*{sec:vggt_attention}
  & \hyperref[sec:vggt_attention]{On 3D Large Reconstruction Model}\dotfill
  & \pageref{sec:vggt_attention} \\[6pt]

\multicolumn{3}{@{}l}{\textbf{Implementation}} \\
\S\,\ref*{sec:CUDA}
  & \hyperref[sec:CUDA]{CUDA C++ Backend and Portability}\dotfill
  & \pageref{sec:CUDA} \\
\S\,\ref*{sec:Variants}
  & \hyperref[sec:Variants]{ELSA Variants: Precision and Memory}\dotfill
  & \pageref{sec:Variants} \\[6pt]

\multicolumn{3}{@{}l}{\textbf{Limitations}} \\
\S\,\ref*{sec:Limitations}
  & \hyperref[sec:Limitations]{Limitations}\dotfill
  & \pageref{sec:Limitations} \\

\end{tabular}

\renewcommand{\thesection}{\Alph{section}}
\setcounter{section}{0}
\section{Monoid Proof for \texorpdfstring{$\oplus$}{⊕}}
\label{sec:Monoid}


ELSA's parallelization strategy rests on casting online softmax attention
as a parallel prefix scan over an associative operator
(Sec.~\ref{sec:methodology} of the main paper).
This section provides the complete proof that our merge operator $\oplus$
forms a monoid, thereby guaranteeing that
\textcolor{cvprblue}{\textbf{any ordering of pairwise merges yields the
same result}}---the property that makes Hillis--Steele and Blelloch scans
provably correct for ELSA.

\subsection{State Space and Operations}

Let $\bar{\mathbb{R}} = \mathbb{R} \cup \{-\infty\}$ denote the extended
reals, and let $b$, $h$, $d$, $d_v$ denote batch size, number of heads,
key/query dimension, and value dimension respectively.
Define the state space
\begin{equation}\label{eq:supp-state-space}
    G = \bar{\mathbb{R}} \times \mathbb{R}_{\ge 0} \times \mathbb{R}^{d_v},
\end{equation}
whose elements are triples $(m, S, W)$: the running maximum logit $m$,
the normalized cumulative sum $S \in \mathbb{R}_{\ge 0}$, and the
weighted value accumulator $W \in \mathbb{R}^{d_v}$.
For $a=(m_a,S_a,W_a)$ and $b=(m_b,S_b,W_b)$ in $G$, we define two
operations that underpin $\oplus$
(Eq.~\eqref{eq:merge-compact} of the main paper).

\noindent\textbf{Unnormalize:}
\begin{equation}\label{eq:supp-unnorm}
    \mathrm{un}(m,S,W) \;=\; (m,\; Se^{m},\; We^{m})
\end{equation}
exposes the raw accumulated sums prior to numerical stabilization,
reversing the $e^{-m}$ re-anchoring applied during the scan.

\noindent\textbf{Renormalize:}
\begin{equation}\label{eq:supp-renorm}
    \mathrm{renorm}(m,\tilde{S},\tilde{W}) \;=\;
    (m,\; \tilde{S}e^{-m},\; \tilde{W}e^{-m})
\end{equation}
restores numerical stability by rescaling with the running maximum,
ensuring all exponents remain $\le 0$ (Lemma~1 of the main paper).

The merge operator $\oplus$ composes two block states via
\textcolor{cvprblue}{\textbf{un-sum-renorm}}:
\begin{equation}\label{eq:supp-merge}
\begin{aligned}
    a \oplus b \;&=\; \mathrm{renorm}\!\left(m_c,\;
      \tilde{S}_a + \tilde{S}_b,\; \tilde{W}_a + \tilde{W}_b\right), \\
    m_c \;&=\; \max(m_a, m_b),
\end{aligned}
\end{equation}
where $(m_a,\tilde{S}_a,\tilde{W}_a) = \mathrm{un}(a)$ and
$(m_b,\tilde{S}_b,\tilde{W}_b) = \mathrm{un}(b)$.

\subsection{Monoid Structure}

\paragraph{Identity element.}
Let $\boldsymbol{e} = (-\infty, 0, \mathbf{0})$.
Since $\mathrm{un}(\boldsymbol{e}) = (-\infty, 0, \mathbf{0})$,
by Eq.~\eqref{eq:supp-merge}, $\boldsymbol{e}$ contributes zero to both
$\tilde{S}$ and $\tilde{W}$, and $m = -\infty$ never dominates any finite
$m$ under $\max$. Hence $\boldsymbol{e}$ is a
\textcolor{cvprblue}{\textbf{two-sided identity}}:
\begin{equation}\label{eq:supp-identity}
    a \oplus \boldsymbol{e} \;=\; \boldsymbol{e} \oplus a \;=\; a
    \qquad \forall\, a \in G,
\end{equation}
which enables Blelloch's down-sweep to initialize boundary states
correctly without special-casing empty prefixes.

\paragraph{Associativity.}
Let $u_a, u_b, u_c \in G$ with $u_i = (m_i, S_i, W_i)$ for
$i \in \{a,b,c\}$, and let $\tilde{S}_i = S_i e^{m_i}$,
$\tilde{W}_i = W_i e^{m_i}$ denote their unnormalized sums.
For $u_{ab} = u_a \oplus u_b$, Eq.~\eqref{eq:supp-merge} gives
$m_{ab} = \max(m_a, m_b)$ and $\tilde{S}_{ab} = \tilde{S}_a + \tilde{S}_b$,
$\tilde{W}_{ab} = \tilde{W}_a + \tilde{W}_b$.
Merging with $u_c$ and writing $m_* = \max(m_a, m_b, m_c)$:
\begin{equation}\label{eq:supp-lhs}
\scalebox{0.88}{$\displaystyle
    S_{(ab)c} = \bigl(\tilde{S}_a + \tilde{S}_b + \tilde{S}_c\bigr)e^{-m_*}, \quad
    W_{(ab)c} = \bigl(\tilde{W}_a + \tilde{W}_b + \tilde{W}_c\bigr)e^{-m_*}.
$}
\end{equation}
The right-associative grouping $u_a \oplus (u_b \oplus u_c)$ yields the
same $m_*$ and the same unnormalized totals, so
\begin{equation}\label{eq:supp-rhs}
\scalebox{0.88}{$\displaystyle
    S_{a(bc)} = \bigl(\tilde{S}_a + \tilde{S}_b + \tilde{S}_c\bigr)e^{-m_*}, \quad
    W_{a(bc)} = \bigl(\tilde{W}_a + \tilde{W}_b + \tilde{W}_c\bigr)e^{-m_*}.
$}
\end{equation}
Since Eqs.~\eqref{eq:supp-lhs} and~\eqref{eq:supp-rhs} are identical,
\begin{equation}\label{eq:supp-assoc}
    (u_a \oplus u_b) \oplus u_c \;=\; u_a \oplus (u_b \oplus u_c).
\end{equation}
Together with Eq.~\eqref{eq:supp-identity}, $(G, \oplus, \boldsymbol{e})$
is a \textcolor{cvprblue}{\textbf{monoid}}, guaranteeing the correctness
of the two-level parallel prefix scan in Sec.~\ref{sec:methodology} and
the FP32 error bound in Theorem~1 of the main paper. \hfill$\square$

\subsection{Derivation of the Floating-Point Error Bound}
\label{sec:error_derivation}

We derive the bound stated in Theorem~1 of the main paper.
Throughout, we adopt the standard IEEE-754 model
$\mathrm{fl}(x \mathbin{\mathrm{op}} y) = (x \mathbin{\mathrm{op}} y)(1+\delta)$
with $|\delta| \le u$ (unit roundoff), and rely on Lemma~1 (bounded
exponents) and Assumption~1 (exp-accuracy) from the main paper.

\paragraph{Per-merge error.}
Consider a single pairwise merge
$S_{\mathrm{out}} = S_a e^{m_a - m_{\mathrm{out}}} +
S_b e^{m_b - m_{\mathrm{out}}}$.
Rounding errors arise from (1)~the exponential scaling factors and
(2)~the subsequent addition.
By Lemma~1, all scaling exponents satisfy $m_a - m_{\mathrm{out}} \le 0$,
so $e^{m_a - m_{\mathrm{out}}} \in (0,1]$---bounding the mantissa and
preventing error amplification.
Letting $\epsilon_k$ denote the relative error at depth $k$ of the
reduction tree, the recursive step satisfies
\begin{equation}\label{eq:supp-merge-err}
    \hat{S}_{k} \;\approx\;
    S_{\mathrm{true}}\,(1 + \epsilon_{k-1})
    (1 + \delta_{\mathrm{exp}})(1 + \delta_{\mathrm{add}}).
\end{equation}
Dropping second-order terms $\mathcal{O}(u^2)$, the error grows
\textcolor{cvprblue}{\textbf{linearly in depth}}:
\begin{equation}\label{eq:supp-per-merge}
    |\epsilon_k| \;\le\; |\epsilon_{k-1}| + c\,u,
\end{equation}
where $c$ counts the FLOPs per merge (one subtraction, one exponential,
one multiplication, one addition).

\paragraph{Total depth and error bound.}
The two-level scan (Sec.~\ref{sec:methodology} of the main paper)
contributes
\begin{equation}\label{eq:supp-depth}
\scalebox{0.88}{$\displaystyle
    L(n, B) \;=\;
    \underbrace{\lceil \log_2 B \rceil}_{\text{intra-block (Hillis--Steele)}}
    +\;
    \underbrace{2\lceil \log_2(n/B) \rceil}_{\text{inter-block (Blelloch up/down)}}
    +\; \mathrm{const}
$}
\end{equation}
matching Eq.~\eqref{eq:depth} of the main paper.
Telescoping Eq.~\eqref{eq:supp-per-merge} over $L(n,B)$ levels gives
\begin{equation}\label{eq:supp-bound}
    \frac{\|\hat{y} - y\|_2}{\|y\|_2}
    \;\le\; \gamma_{L(n,B)} \;\approx\; L(n,B)\cdot u,
\end{equation}
where $\gamma_n = nu/(1-nu) \approx nu$ (Higham~\cite{Higham2002Accuracy}).
Since \textcolor{cvprblue}{\textbf{$L(n,B) = \mathcal{O}(\log n)$}}, the
error grows logarithmically with sequence length---in contrast to naive
summation ($\mathcal{O}(\sqrt{n}\,u)$) or sequential recurrence
($\mathcal{O}(n\,u)$)---confirming the numerical stability of ELSA as
$n \to \infty$ and establishing Theorem~1.

\subsection{Algorithm}
\label{sec:algA4}

This subsection provides the full pseudocode and a step-by-step
walk-through of the per-query ELSA scan, which was condensed in the main
paper (Sec.~\ref{sec:methodology}) due to space constraints.
Algorithm~\ref{alg:scan} writes the scan in its sequential single-query
form for clarity; the parallel two-level realization on GPU follows
directly from the monoid structure of $\oplus$ and is described in
Sec.~\ref{sec:CUDA}.

\paragraph{State triple and invariants.}
The algorithm maintains a triple $(m_j, S_j, W_j)$ summarizing the prefix
$1{:}j$ of the key/value stream, where $m_j$ is the running maximum logit,
$S_j=\sum_{i\le j}\exp(s_i-m_j)$ is the rescaled normalizer, and
$W_j=\sum_{i\le j}\exp(s_i-m_j)\,V[i]$ is the rescaled weighted value sum.
These three quantities together encode the exact softmax-weighted output
for the prefix as $W_j/S_j$, while remaining numerically stable because
all exponentials are evaluated relative to the current maximum $m_j$.
The identity element $\boldsymbol{e}=(-\infty,0,\mathbf{0})$ initializes
the recursion (Line~3) so that the first iteration encounters
$m_0=-\infty$ and immediately enters the ``new max'' branch.

\paragraph{Per-token update.}
For each new token $j$ (Lines~4--16), the algorithm first computes the
scaled inner product
$s_j=\langle q,K[j]\rangle/\sqrt{d}$ (Line~5).
It then dispatches on whether the previous running maximum still
dominates ($m_{j-1}\ge s_j$, Lines~6--10) or the new logit takes over
($m_{j-1}<s_j$, Lines~11--15):
\begin{itemize}[leftmargin=12pt,topsep=2pt,itemsep=0pt]
  \item \textbf{Case A: maximum unchanged.} The previous $m_{j-1}$ is
  preserved as $m_j$ (Line~7). The new contribution is folded in by
  computing $\alpha=\exp(s_j-m_j)$ (Line~8) and adding it to both the
  normalizer $S$ and the weighted value sum $W$
  (Lines~9--10). No rescaling of past terms is required because the
  normalization anchor $m_j$ has not moved.
  \item \textbf{Case B: new maximum.} The current logit $s_j$ becomes the
  new anchor (Line~12). All previously accumulated quantities are
  rescaled by $\beta=\exp(m_{j-1}-m_j)\le 1$ (Line~13) so that they are
  expressed relative to the new maximum, and the contribution of token
  $j$ itself is added with weight $\exp(0)=1$ (Lines~14--15).
\end{itemize}
This conditional rescaling is precisely the online-softmax update of
Eq.~\eqref{eq:online-softmax}: it guarantees that, after every step,
$(m_j,S_j,W_j)$ encodes the exact softmax of the prefix without ever
materializing the $n\times n$ score matrix and without floating-point
overflow, since all exponentials use non-positive arguments.

\paragraph{Final normalization.}
After processing all $n$ tokens, the attention output is recovered by a
single division $y=W_n/S_n$ (Line~17). Because $W_n$ and $S_n$ are
expressed in the same exponential frame anchored at $m_n$, the
$\exp(\cdot)$ scale cancels in the ratio, yielding the exact softmax
output up to floating-point round-off bounded by $u\,L(n,B)$
(Theorem~\ref{thm:fp32} of the main paper).

\paragraph{From sequential to parallel.}
Although Algorithm~\ref{alg:scan} is presented as a left-to-right loop
for readability, its correctness depends only on the associativity of
$\oplus$ (Proposition~\ref{prop:block}), not on the order of
combination. Concretely, any binary tree of $\oplus$-merges over the
state triples $u_1,\dots,u_n$ produces the identical final state
$u_1\oplus\cdots\oplus u_n$. This is exactly the property exploited by
the GPU implementation: intra-block reductions use a Hillis--Steele tree
of depth $\lceil\log_2 B\rceil$ in shared memory, and inter-block
combinations use a Blelloch up-/down-sweep of depth
$2\lceil\log_2(n/B)\rceil$ in global memory, yielding a total scan depth
$L(n,B)$ (Sec.~\ref{sec:CUDA}). The pseudocode below should therefore
be read as a \emph{specification} of $\oplus$ applied left-to-right; any
parallel evaluation order matching the same monoid contract produces
bit-equivalent results in exact arithmetic and bounded round-off in
floating point.

\begin{algorithm}[tb]
\caption{ELSA per-query sequential formulation. The loop is written
sequentially for clarity; parallel execution follows from the monoid
structure of $\oplus$ (Sec.~\ref{sec:methodology}).}
\label{alg:scan}
\scalebox{0.70}[0.70]{
\begin{minipage}{1.41\linewidth}
\begin{algorithmic}[1]
\Require Query $q\in \mathbb{R}^d$, Keys $K[1..n] \in (\mathbb{R}^d)^n$, Values $V[1..n] \in (\mathbb{R}^{d_v})^n$
\Ensure Attention output $y \in \mathbb{R}^{d_v}$
\State $(m_0, S_0, W_0) \gets (-\infty, 0, \mathbf{0})$\Comment{initialize state triple: max logit, normalized sum, weighted sum}
\For{$j = 1$ to $n$}
\State $s_j \gets \frac{\langle q, K[j]\rangle}{\sqrt{d}}$\Comment{compute attention score (logit) for token $j$}
\If{$m_{j-1} \ge s_j$}
\Comment{current max remains; rescale new contribution}
\State $m_j \gets m_{j-1}$\Comment{keep previous maximum}
\State $\alpha \gets \exp(s_j - m_j)$\Comment{normalized weight for token $j$}
\State $S_j \gets S_{j-1} + \alpha$\Comment{accumulate normalized sum}
\State $W_j \gets W_{j-1} + \alpha V[j]$\Comment{accumulate weighted value sum}
\Else
\Comment{new max found; rescale all previous contributions}
\State $m_j \gets s_j$\Comment{update maximum to current score}
\State $\beta \gets \exp(m_{j-1} - m_j)$\Comment{rescaling factor for previous terms}
\State $S_j \gets S_{j-1} \cdot \beta + 1$\Comment{rescale previous sum and add $\exp(0)=1$}
\State $W_j \gets W_{j-1} \cdot \beta + V[j]$\Comment{rescale previous weighted sum and add current value}
\EndIf
\EndFor
\State $y \gets W_n / S_n$\Comment{normalize: divide accumulated weighted sum by total weight}
\end{algorithmic}
\end{minipage}
}
\end{algorithm}

%
\section{I/O Analysis and Bounds}
\label{sec:IO}

We analyze ELSA's I/O complexity at two levels, complementing the I/O
scope paragraph in Sec.~\ref{sec:methodology} of the main paper.
Throughout, \emph{linear-scan} refers to $\mathcal{O}(n)$ extra memory
and one-pass I/O per query; total arithmetic work remains $\mathcal{O}(n^2)$.

\subsection{Per-Query Streaming}

Any exact attention computation for a fixed query must read its $K, V$
inputs and write its output at least once, incurring $\Omega(n)$ I/O.
ELSA \textcolor{cvprblue}{\textbf{attains this lower bound}} by streaming
$K$ and $V$ exactly once per query while keeping the running state
$(m, S, W)$ in registers or fast memory---a direct consequence of the
monoid structure: the identity $\boldsymbol{e}=(-\infty,0,\mathbf{0})$
allows the scan to initialize without reading any additional data.

\subsection{All-Queries Tiling}

For the full $QK^\top$ interaction and subsequent $PV$ accumulation,
classical red--blue pebble arguments imply the I/O lower bound
\begin{equation}\label{eq:io-lower}
    \Omega\!\left(\frac{n^2(d+d_v)}{\sqrt{M_f}}\right)
\end{equation}
for fast-memory capacity $M_f$.
ELSA adopts a blocked schedule with query tile $T_q$ and key/value tile
$T_k$ subject to
\begin{equation}\label{eq:tile-constraint}
    T_q d + T_k(d+d_v) \;\le\; c\,M_f
\end{equation}
for a small constant $c$. The resulting total DRAM traffic is
\begin{equation}\label{eq:dram-traffic}
\scalebox{0.88}{$\displaystyle
    \Theta(nd + nd_v) \;+\; \Theta\!\left(\frac{n^2(d+d_v)}{T_q}\right).
$}
\end{equation}
Setting $T_q \asymp \sqrt{M_f/(d+d_v)}$ reduces this to
\begin{equation}\label{eq:io-upper}
\scalebox{0.88}{$\displaystyle
    \mathcal{O}\!\left(\frac{n^2(d+d_v)}{\sqrt{M_f}}\right) + \Theta(nd + nd_v),
$}
\end{equation}
\textcolor{cvprblue}{\textbf{matching the lower bound}}~\eqref{eq:io-lower}
up to constants (Eq.~\eqref{eq:dram} of the main paper).
Hence \emph{linear-scan} refers strictly to per-query memory
($\mathcal{O}(n)$); the global all-queries schedule meets the
rectangular-matrix I/O lower bound within constant factors via tiling.

\section{Numerical Equivalence Verification}
\label{sec:VecEq}

Standard softmax attention materializes the full $n\times n$ score matrix
and normalizes in a single pass---a numerically well-characterized
reference implementation.
\citet{milakov2018online} showed that the same result can be obtained by
an \emph{online} procedure that maintains running statistics
$(m_i, s_i, \mathbf{w}_i)$ and updates them token by token, without ever
storing the full matrix.
ELSA lifts this sequential scan to a parallel prefix computation by
casting those statistics as elements of an associative monoid
(Sec.~\ref{sec:methodology}); because the merge operator $\oplus$ is
associative, any evaluation order---sequential or parallel---yields
identical results in exact arithmetic.
To validate this claim empirically, we use the vectorized batch
formulation as a verification oracle: it computes $(\hat{P}, \hat{Y})$
via standard matrix operations and we measure the drift against the
monoid scan on GPU.
In floating point the two agree to within $\mathcal{O}(u\log n)$
(Sec.~\ref{sec:error_derivation}), with both sharing the same asymptotic
cost---$\mathcal{O}(bhnd)$ for the query--key product and
$\mathcal{O}(bhnd_v)$ for value accumulation.

\subsection{Vectorized Reference Algorithm}

Let $Q, K \in \mathbb{R}^{b\times h\times n\times d}$ and
$V \in \mathbb{R}^{b\times h\times n\times d_v}$ denote queries, keys,
and values with batch size $b$, $h$ heads, and sequence length $n$.
The reference procedure computes attention probabilities $\hat{P}$ and
output $\hat{Y}$ as
\begin{equation}\label{eq:vec-all}
\scalebox{0.88}{$\displaystyle
\begin{aligned}
    \mathbf{S} &= \tfrac{1}{\sqrt{d}}\,QK^{\top}, &
    m_i &= \max_{j}\,\mathbf{S}_{:,:,i,j}, \\[2pt]
    E   &= \exp(\mathbf{S} - m), &
    s_i &= \textstyle\sum_{j} E_{:,:,i,j}, \\[2pt]
    \hat{P} &= E \oslash s, &
    \hat{Y} &= (EV) \oslash s,
\end{aligned}
$}
\end{equation}
where $\oslash$ denotes row-wise division.
This mirrors the accumulation rules of Algorithm~\ref{alg:scan}:
row-wise max subtraction for stability, exponentiation, sum
accumulation, and normalization.
Numerical drift relative to the scan is reported in
Sec.~\ref{sec:numerical-drift}.

\begin{algorithm}[t]
\caption{Vectorized equivalence check}
\label{alg:vec-eq}
\small
\begin{algorithmic}[1]
\Require $Q, K, V \in \mathbb{R}^{b \times h \times n \times d}$;\;
         reference outputs $P_{\mathrm{ref}},\,Y_{\mathrm{ref}}$
         (high precision)
\Ensure  Drift metrics
         $\bigl\{\|\delta P\|_\infty,\,\|\delta Y\|_\infty,\,\ldots\bigr\}$
\State $\mathbf{S} \gets (QK^\top)/\sqrt{d}$
\State $m \gets \max(\mathbf{S},\;\mathrm{dim}{=}{-1},\;
       \mathrm{keepdim}{=}\mathrm{True})$
\State $E \gets \exp(\mathbf{S} - m)$
\State $s \gets \textstyle\sum(E,\;\mathrm{dim}{=}{-1},\;
       \mathrm{keepdim}{=}\mathrm{True})$
\State $\hat{P} \gets E \oslash s$;\quad $\hat{Y} \gets (EV) \oslash s$
\State \Return $\|\hat{P} - P_{\mathrm{ref}}\|_\infty$,\;
       $\|\hat{Y} - Y_{\mathrm{ref}}\|_\infty$
\end{algorithmic}
\end{algorithm}
\subsection{Evaluation Results and Analysis}
\label{sec:numerical-drift}
\paragraph{Metrics and Setup.}
We define six complementary drift metrics on
$\delta P = \hat{P} - P_{\mathrm{ref}}$ and
$\delta Y = \hat{Y} - Y_{\mathrm{ref}}$,
chosen to probe different failure modes of a numerical reformulation:

\begin{enumerate}[leftmargin=*, itemsep=2pt]
  \item \textbf{Maximum absolute probability drift}\;
        $\|\delta P\|_\infty = \max_{b,h,i,j}|\delta P_{b,h,i,j}|$:
        worst-case pointwise error in the attention matrix---detects any
        single catastrophically mis-computed weight.

  \item \textbf{Relative $L_2$ probability error}\;
        $\|\delta P\|_2/\|P_{\mathrm{ref}}\|_2$:
        aggregate error normalized by scale---captures global distributional
        shift that pointwise maxima may mask by averaging over all entries.

  \item \textbf{Mean Jensen--Shannon divergence}\;
        $\mathrm{JS}_{\mathrm{mean}} = \frac{1}{bhn}\sum_{b,h,i}
        \mathrm{JS}(\hat{P}_{b,h,i,:}\,\|\,P_{\mathrm{ref},b,h,i,:})$:
        per-row distributional discrepancy---JSD is symmetric and bounded,
        so it measures whether the \emph{shape} of each query's attention
        distribution is preserved, beyond individual entry magnitudes.

  \item \textbf{Argmax disagreement rate}\;
        $\mathrm{argRate} = \frac{1}{bhn}\sum_{b,h,i}
        \mathbf{1}[\arg\max_j\hat{P}_{b,h,i,j}
        \neq\arg\max_j P_{\mathrm{ref},b,h,i,j}]$:
        fraction of query positions where the top-attended token changes---
        the most behaviorally critical metric, as argmax disagreement
        directly alters which context the model attends to.

  \item \textbf{Maximum absolute output drift}\;
        $\|\delta Y\|_\infty = \max_{b,h,i,k}|\delta Y_{b,h,i,k}|$:
        worst-case error in the final output embeddings propagated to the
        next layer---the downstream analogue of metric~1.

  \item \textbf{Relative $L_2$ output error}\;
        $\|\delta Y\|_2/\|Y_{\mathrm{ref}}\|_2$:
        normalized aggregate output error---confirms that overall output
        magnitude is faithfully preserved across the full tensor.
\end{enumerate}

Metrics 1--4 probe the attention probability matrix $P$, progressing
from pointwise to distributional to functional correctness;
metrics 5--6 repeat the same hierarchy for the output $Y$.
All are computed in a fully vectorized manner on GPU; we report
the median, 95th, and 99th percentiles across all tokens and heads.

\paragraph{Results.}

Table~\ref{tab:eq-p95} reports 95th-percentile drift between the monoid
scan (FP32) and a float64 oracle reference across all scenarios.
\textcolor{cvprblue}{\textbf{Discrepancies reach at most $3.3\times10^{-16}$
for probabilities and $3.3\times10^{-15}$ for outputs}}---well below
single-precision unit roundoff $u \approx 6\times10^{-8}$, confirming
that the scan result is indistinguishable from the float64 reference at
FP32 precision.
Crucially, the argmax disagreement rate is identically zero in all
scenarios, confirming that the monoid scan \textcolor{cvprblue}{\textbf{exactly
preserves the top attention weight}}.
Long-sequence and stress scenarios exhibit slightly larger absolute errors
owing to accumulated partial sums, yet all values remain at
float64-oracle-level drift---as predicted by the logarithmic depth
$L(n,B)$ (Eq.~\eqref{eq:supp-depth}).

\begin{figure}[t]
  \centering
  \includegraphics[width=0.48\linewidth]{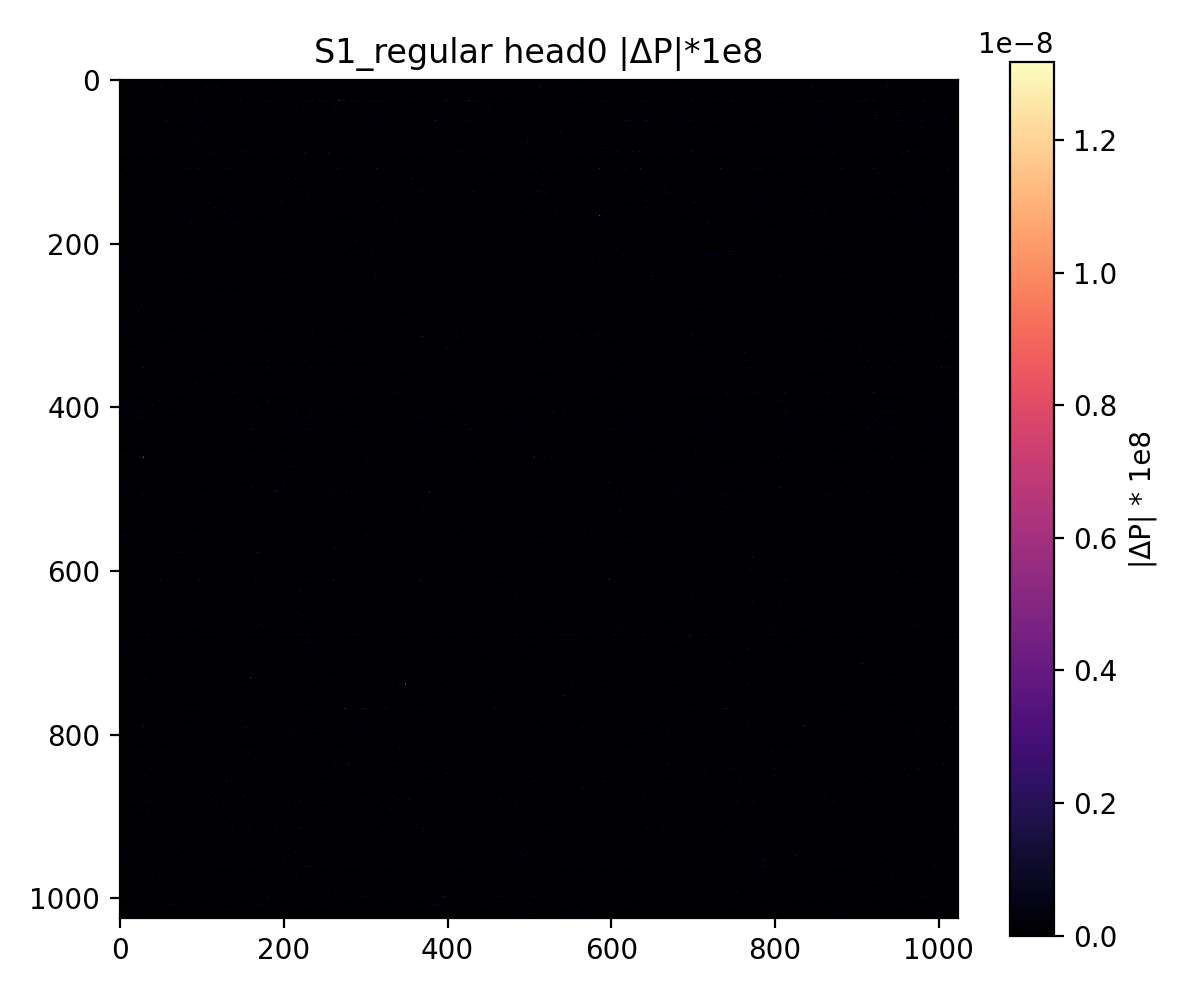}
  \hfill
  \includegraphics[width=0.48\linewidth]{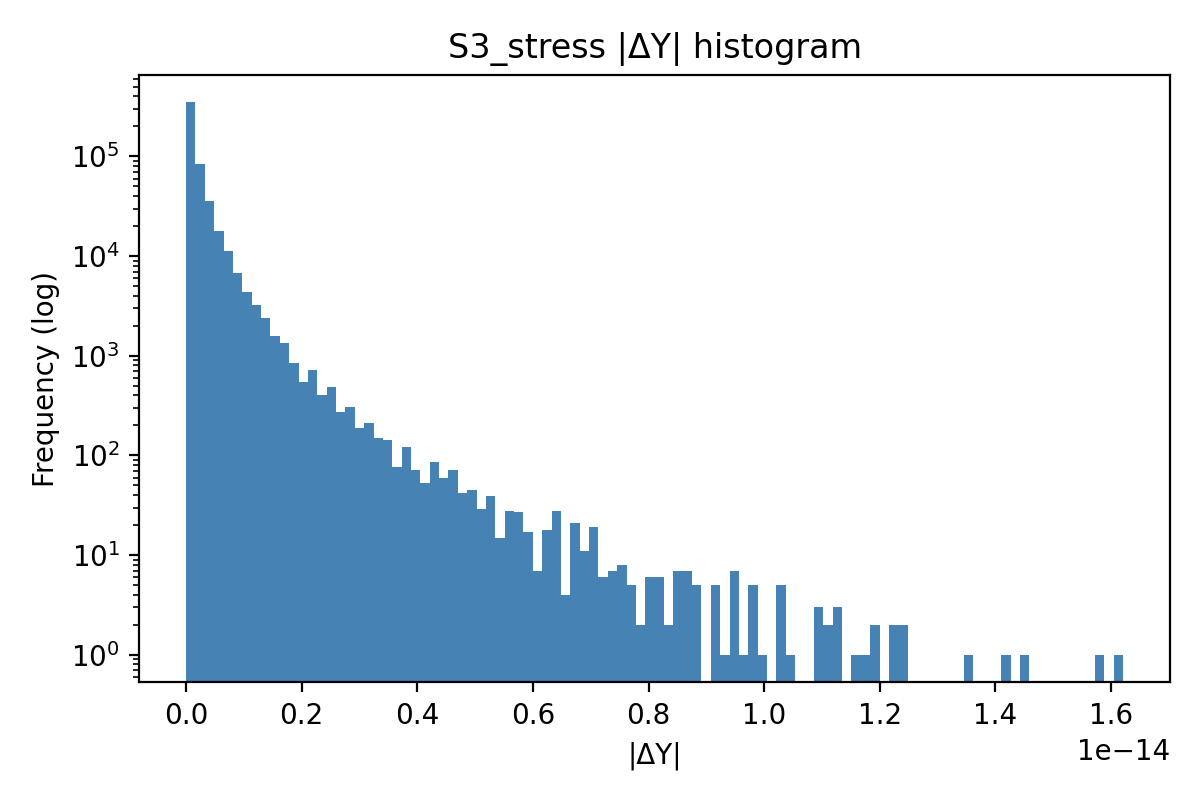}\caption{\textbf{Drift visualization.}
  Left: heatmap of $|\delta P|$ (scaled by $10^8$) for head~0, regular
  scenario; the uniformly dark field indicates no systematic spatial
  accumulation, errors are random floating-point noise throughout the
  $n{\times}n$ attention matrix.
  Right: log-scale histogram of $|\delta Y|$ under the stress scenario;
  the distribution is heavily concentrated near zero, with the tail
  bounded below $2\times10^{-14}$, confirming machine-precision agreement
  in the worst case.}

  \label{fig:eq-heat-hist}
\end{figure}

\begin{figure}[t]
  \centering
  \includegraphics[width=\linewidth]{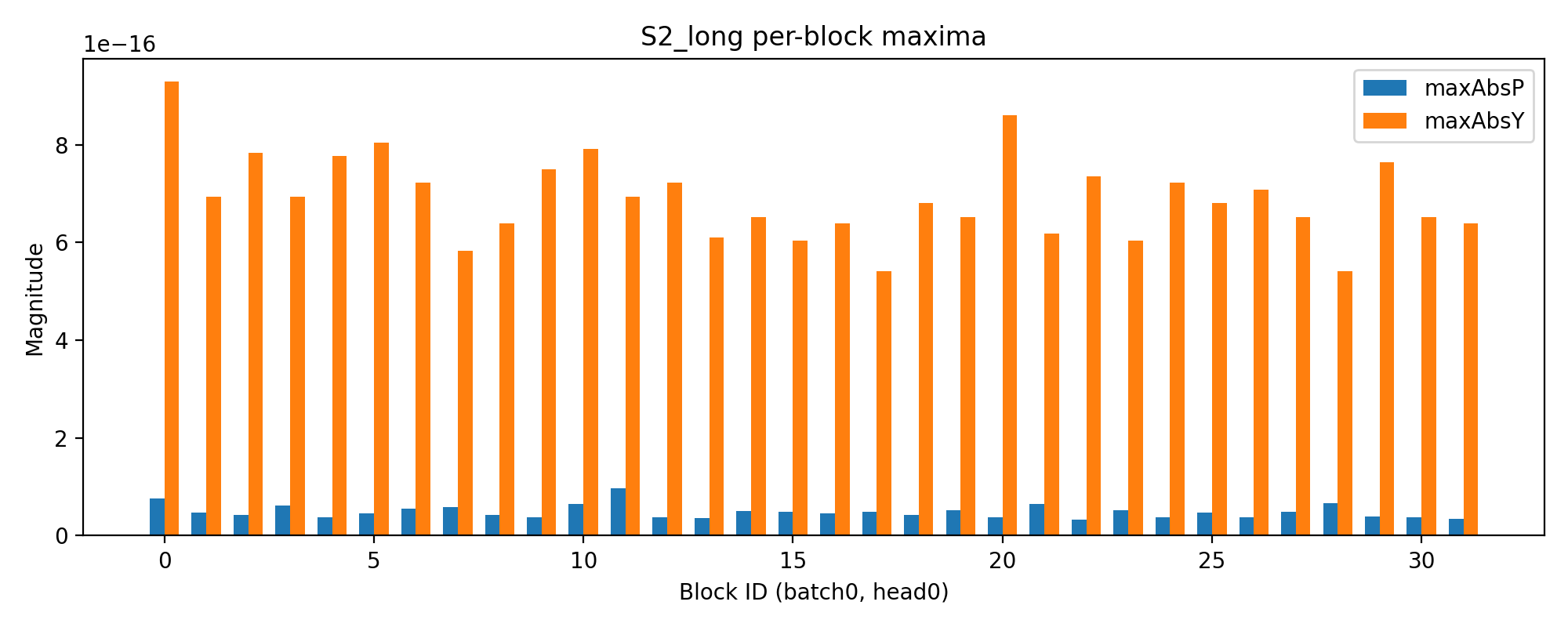}
  \caption{\textbf{Blockwise monoid validation.}
    Tokens are partitioned into blocks of 128; $(m,S,W)$ is computed per block in parallel
    and then reduced via $\oplus$. The resulting drift metrics are indistinguishable from the
    global vectorized procedure, validating correctness of the associative merge.}
  \label{fig:eq-block}
\end{figure}

Figure~\ref{fig:eq-heat-hist} further corroborates these findings:
the heatmap (left) reveals no spatial structure in $|\delta P|$ across
token pairs---the uniformly dark field confirms that errors are random
floating-point noise rather than systematic accumulation---while the
log-scale histogram (right) shows $|\delta Y|$ heavily concentrated
near zero even under the stress scenario, with its tail bounded well
below $2\times10^{-14}$, orders of magnitude beneath single-precision
unit roundoff.
Finally, Figure~\ref{fig:eq-block} directly validates monoid composition
(Proposition~1) by partitioning tokens into blocks of $B=128$, computing
$(m,S,W)$ per block in parallel, and reducing via $\oplus$; the resulting
per-block drift is indistinguishable from the float64 oracle result,
confirming correctness of the associative merge at the core of ELSA.

\begin{table}[t]
  \centering
  \small
  \setlength{\tabcolsep}{4pt}
  \begin{tabular}{lccc}
    \toprule
    \textbf{Metric (p95)} & \textbf{Regular} & \textbf{Long} & \textbf{Stress} \\
    \midrule
    $\|\delta P\|_\infty$                       & $3.12\times10^{-17}$ & $2.34\times10^{-17}$ & $3.33\times10^{-16}$ \\
    $\|\delta P\|_2/\|P_{\mathrm{ref}}\|_2$     & $1.73\times10^{-15}$ & $3.42\times10^{-15}$ & $3.50\times10^{-15}$ \\
    $\mathrm{JS}_{\mathrm{mean}}$               & $3.56\times10^{-16}$ & $3.77\times10^{-16}$ & $3.07\times10^{-16}$ \\
    $\mathrm{argRate}$                          & $0$                  & $0$                  & $0$                  \\
    $\|\delta Y\|_\infty$                       & $4.99\times10^{-16}$ & $4.99\times10^{-16}$ & $3.28\times10^{-15}$ \\
    $\|\delta Y\|_2/\|Y_{\mathrm{ref}}\|_2$     & $2.39\times10^{-15}$ & $4.72\times10^{-15}$ & $4.94\times10^{-15}$ \\
    \bottomrule
  \end{tabular}
  \caption{\textbf{Vectorized equivalence (oracle-based).} 95th-percentile drift metrics against a high-precision (float64) reference, across all batches,
    heads, and tokens. All three scenarios show near-machine-precision agreement.}
  \label{tab:eq-p95}
\end{table}

\section{Ablation Studies}
\label{sec:ablation}

\begin{figure}[t]
  \centering
  \includegraphics[width=0.9\linewidth]{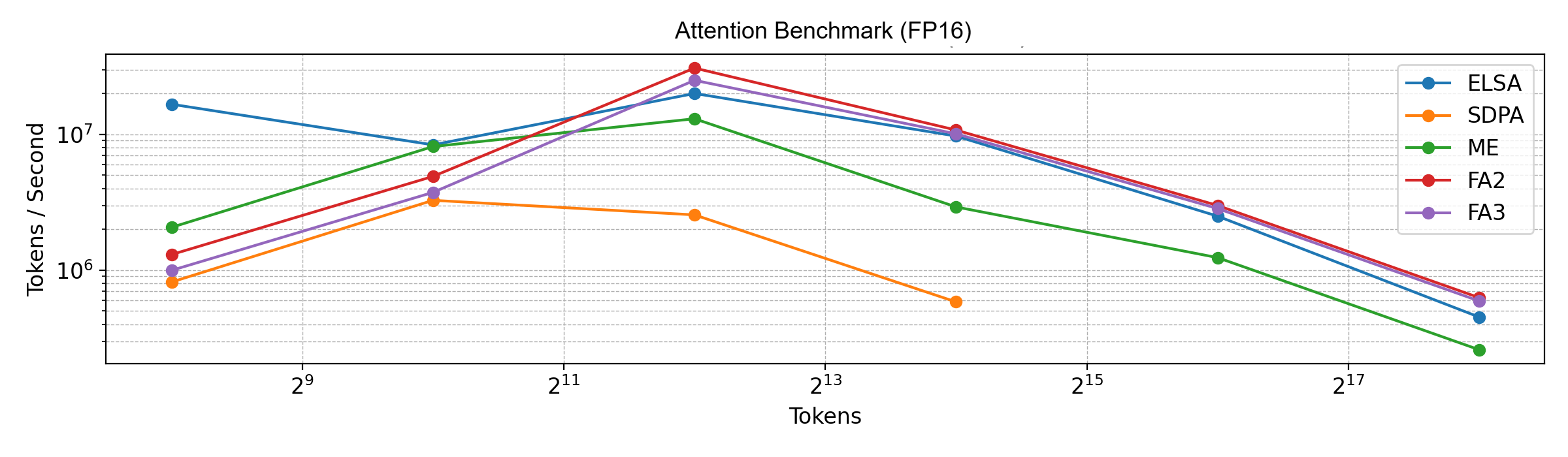}
  \vspace{-2pt}
  \includegraphics[width=0.9\linewidth]{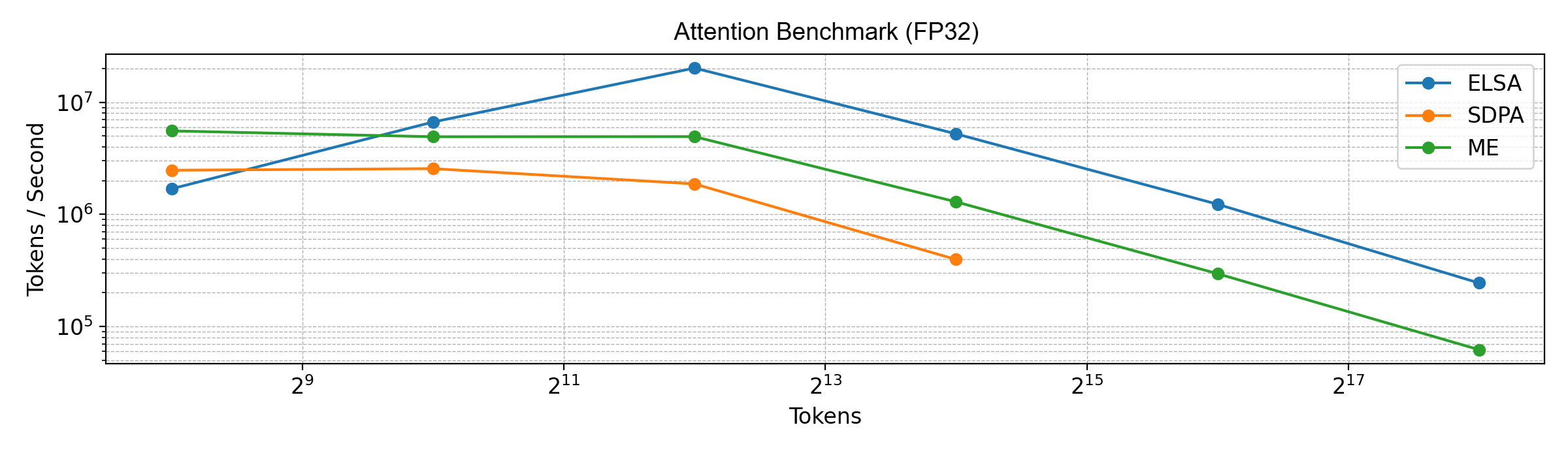}
  \vspace{-6pt}
  \caption{\textbf{Memory footprint on synthetic benchmarks.} ELSA's advantage widens with sequence length in both precision modes.}
  \label{fig:memory_fp}
  \vspace{-8pt}
\end{figure}

We examine the impact of block size ($B$), accumulator precision, and
resource-constrained deployment (Fig.~\ref{fig:memory_fp}).
Block size $B{=}128$ offers the best latency-occupancy trade-off across
all configurations.
Using FP16 for the $(S,W)$ accumulator causes numerical error to increase
dramatically beyond 2K tokens, while FP32 remains stable throughout
(Fig.~\ref{fig:memory_fp}, bottom).
On Jetson TX2, baseline MHSA kernels and FA2 cannot process long sequences
due to memory limitations, whereas ELSA maintains efficiency across all
token lengths (Fig.~\ref{fig:memory_fp}, top).

\section{Variable-Length Sequence Benchmark}
\label{sec:varlen}

To ensure that performance differences are not confounded by data-layout conversions, we
benchmark variable-length sequences. We generate 8 sequences with lengths uniformly sampled
from $[256, 1792]$ and compare ELSA (direct processing) against FA2 in unpadded mode
(\texttt{cu\_seqlens}), accounting for FA2's conversion cost between padded and packed layouts.

Table~\ref{tab:varlen} reports end-to-end latency, kernel time, conversion overhead, peak
memory, and throughput (medians over 200 runs). ELSA's zero-copy pipeline reduces FA2's padded-to-packed conversion overhead (from $1.34$\,ms to $0.66$\,ms), achieving comparable end-to-end latency despite a
longer kernel time. 

\begin{table}[t]
\centering
\small
\scalebox{0.88}{%
\setlength{\tabcolsep}{2pt}
\begin{tabular}{lccccc}
\toprule
\textbf{Method} & \textbf{Total (ms)} & \textbf{Kernel (ms)} & \textbf{Convert (ms)} & \textbf{Mem (GB)} & \textbf{Tput (M)} \\
\midrule
ELSA & 1.79 & 1.12 & \cellcolor{bestgray}{\textbf{0.66}} & \cellcolor{bestgray}{\textbf{0.05}} & 5.66 \\
FA2  & \cellcolor{bestgray}{\textbf{1.65}} & \cellcolor{bestgray}{\textbf{0.31}} & 1.34 & 0.10 & \cellcolor{bestgray}{\textbf{6.14}} \\
\bottomrule
\end{tabular}}
\caption{\textbf{Variable-length sequence benchmark.}
  Random sequences sampled from $[256, 1792]$ (FP32, medians over 200 runs).
  Total: end-to-end latency; Kernel: attention kernel time; Convert: layout conversion
  overhead; Mem: peak GPU memory; Tput: throughput in millions of tokens/s.
  ELSA's zero-copy design reduces conversion overhead at the cost of longer kernel time.}
\label{tab:varlen}
\end{table}

\section{Window Attention and Offloading Analysis}
\label{sec:window_att}

\subsection{Block Size Sensitivity}

Window-ELSA introduces a block size $B$ controlling the work assigned to
each prefix-scan step. We swept $B \in \{64, 128, 256\}$ on Swin-T and
Swin-S (A100, FP32) to characterize the throughput-memory trade-off.
Smaller blocks reduce buffer pressure at the cost of more scan steps;
larger blocks amortize scan overhead but may spill to slower memory.
Concretely, $B{=}64$ reduces throughput by ${\sim}15\%$ while cutting
memory by ${\sim}20\%$; $B{=}256$ yields negligible speedup but increases
memory by ${\sim}20\%$.
Relaxing synchronization to every two blocks saves ${\sim}5\%$ runtime at
a cost of ${\sim}8\%$ extra memory.
We therefore adopt $B{=}128$ with per-block synchronization as the default
operating point; memory-constrained platforms may prefer $B{=}64$.

\subsection{Offloading Performance}

\paragraph{Methodology.}
We quantify PCIe--compute overlap using LLaMA-13B instantiated as a
meta-graph with no materialized weights; layer weights are loaded one
layer at a time during the forward pass via asynchronous double-buffering
(layer $n{+}1$ transfers while layer $n$ computes).
Each layer is instrumented with three CUDA event pairs measuring
(i)~H2D transfer time (\texttt{copy\_ms}), (ii)~kernel computation time
(\texttt{compute\_ms}), and (iii)~idle time (\texttt{wait\_ms}).
We summarize overlap via
\begin{equation}
  \mathrm{Overlap}_{\mathrm{eff}}
    = 1 - \frac{\texttt{wait\_ms}}{\texttt{total\_ms}},
  \label{eq:overlap-eff}
\end{equation}
which approaches 1.0 when PCIe transfers are fully hidden by computation.

\paragraph{Results.}
Table~\ref{tab:offload_filled} reports results for LLaMA-13B (FP32,
batch~1) across four sequence lengths.
At 4K and 8K tokens the workload is PCIe-bound: transfers complete faster
than ELSA's scan overhead, yielding higher idle time (0.727\,ms and
1.005\,ms vs.\ 0.228\,ms and 0.322\,ms for SDPA) and lower throughput
($-$42.4\% and $-$25.2\% respectively).
At 32K tokens, ELSA's lower memory footprint allows computation to begin
earlier, reducing idle time from 540\,ms to 467\,ms/layer and delivering
a \textcolor{cvprblue}{\textbf{17.8\%}} throughput gain (1.460 vs.\
1.239\,M\,tok/s).
At 65K tokens ELSA sustains \textcolor{cvprblue}{\textbf{20.2\%}} higher
throughput (0.882 vs.\ 0.734\,M\,tok/s), confirming that memory efficiency
compounds into end-to-end gains as sequences grow.

\section{On 3D Large Reconstruction Model}
\label{sec:vggt_attention}

To validate ELSA's effectiveness on 3D vision tasks, we evaluate it within the Visual Geometry-Grounded Transformer (VGGT)~\cite{wang2025vggt}, a feed-forward Large Reconstruction Model (LRM) for multi-view 3D reconstruction. VGGT alternates between frame-wise and global attention layers; the global attention aggregates cross-view information but becomes the computational bottleneck as the number of frames grows.

\paragraph{Experimental Setup.}
We compare ELSA against FA2-FP16, xFormers-FP32 (ME-SDPA), and PyTorch Math under two configurations: (1)~\textbf{VGGT Baseline} on 50–150 frames, and (2)~\textbf{FastVGGT}~\cite{shen2025fastvggt}, a memory-optimized variant with token merging. All experiments use $350\times518$ input resolution, yielding 1,041 tokens per frame ($28\times37$ patches plus camera and register tokens).

\paragraph{Results on VGGT Baseline.}
Table~\ref{tab:vggt_baseline} shows that ELSA-FP32 achieves a \textbf{1.46$\times$ speedup} over xFormers at 50 frames (12.56\,s vs.\ 18.36\,s) while matching memory usage (15.74\,GB, within 0.01\,GB). The advantage compounds with sequence length, reaching \textbf{2.34$\times$} at 150 frames (55.97\,s vs.\ 130.83\,s) with an identical 37.59\,GB footprint. FA2-FP16 remains the fastest baseline (3.67\,s at 50 frames) but consumes 17.16\,GB—9\% more than ELSA-FP32.

\paragraph{Results on FastVGGT Scaling.}
Table~\ref{tab:fastvggt} demonstrates ELSA's scalability on longer sequences. At 400 frames, ELSA-FP32 completes inference in 158.04\,s versus xFormers' 218.30\,s (\textbf{1.38$\times$ speedup}), with identical 32.70\,GB memory. The Math baseline runs out of memory beyond 30 frames due to its $O(n^2)$ attention matrix materialization. Across all frame counts, ELSA's memory overhead relative to xFormers remains negligible ($<$0.01\,GB), confirming that the $O(n)$ auxiliary state $(m, S, W)$ imposes no practical burden at these scales.

\paragraph{Analysis.}
ELSA's $O(\log n)$ scan depth confers two key advantages for multi-view 3D reconstruction: (1)~efficient parallelization over long token sequences (10K–100K tokens across views), and (2)~reduced global synchronization overhead compared to xFormers' $O(n/T_k)$ sequential tiling. At 400 frames with 1,041 tokens per frame (416,400 tokens total), the number of global synchronization points drops from ${\sim}3{,}250$ (xFormers, $T_k=128$) to ${\sim}15$ (scan-tree depth $\lceil\log_2(400\!\times\!1041/128)\rceil+3\approx15$), directly explaining the observed 1.38$\times$ end-to-end speedup.

\definecolor{ForestGreen}{rgb}{0.13,0.55,0.13}

\begin{table}[t]
\centering
\small
\setlength{\tabcolsep}{3.5pt}
\begin{tabular}{lcccc}
\toprule
\textbf{Method} & \textbf{Frames} & \textbf{Time (s)} & \textbf{Mem (GB)} & \textbf{Speedup} \\
\midrule
\multirow[c]{3}{*}{xFormers-FP32}
  & 50  & 18.36  & 15.74 & $1.00\times$ \\
  & 100 & 61.75  & 26.67 & $1.00\times$ \\
  & 150 & 130.83 & 37.59 & $1.00\times$ \\
\midrule
\multirow[c]{3}{*}{FA2-FP16}
  & 50  & \textbf{3.67}  & 17.16 & {--} \\
  & 100 & \textbf{9.44}  & 27.13 & {--} \\
  & 150 & 17.88          & 37.43 & {--} \\
\midrule
\multirow[c]{3}{*}{\textbf{ELSA-FP32}}
  & \cellcolor{gray!10}50
  & \cellcolor{gray!10}\textbf{12.56}
  & \cellcolor{gray!10}\textbf{15.74}
  & \cellcolor{gray!10}\textcolor{ForestGreen}{\textbf{$1.46\times$}} \\
  & \cellcolor{gray!20}100
  & \cellcolor{gray!20}\textbf{29.49}
  & \cellcolor{gray!20}\textbf{26.68}
  & \cellcolor{gray!20}\textcolor{ForestGreen}{\textbf{$2.09\times$}} \\
  & \cellcolor{gray!30}150
  & \cellcolor{gray!30}\textbf{55.97}
  & \cellcolor{gray!30}\textbf{37.59}
  & \cellcolor{gray!30}\textcolor{ForestGreen}{\textbf{$2.34\times$}} \\
\bottomrule
\end{tabular}
\caption{\textbf{VGGT baseline.} ELSA-FP32 achieves $1.46$–$2.34\times$
  speedup over xFormers-FP32 while matching its memory footprint.
  Speedup is relative to xFormers-FP32;
  shading intensity reflects speedup magnitude.}
\label{tab:vggt_baseline}
    \vspace{-3mm}
\end{table}

\begin{table}[t]
\centering
\small
\setlength{\tabcolsep}{3pt}
\begin{tabular}{lcccc}
\toprule
\textbf{Method} & \textbf{Frames} & \textbf{Time (s)} & \textbf{Mem (GB)} & \textbf{Speedup} \\
\midrule
\multirow[c]{8}{*}{FA2-FP16}
  & 15  & 1.76  & 4.42  & {--} \\
  & 30  & 2.38  & 4.42  & {--} \\
  & 50  & 3.35  & 5.04  & {--} \\
  & 100 & 6.26  & 7.85  & {--} \\
  & 150 & 9.97  & 10.66 & {--} \\
  & 200 & 14.75 & 13.47 & {--} \\
  & 300 & 27.08 & 19.09 & {--} \\
  & 400 & 46.14 & 24.72 & {--} \\
\midrule
\multirow[c]{3}{*}{Math-FP32}
  & 15    & 4.93  & 8.58  & \textcolor{red!70!black}{$0.68\times$} \\
  & 30    & 13.06 & 18.08 & \textcolor{red!70!black}{$0.45\times$} \\
  & $>$30 & \multicolumn{3}{c}{\textcolor{red!70!black}{\textit{OOM}}} \\
\midrule
\multirow[c]{3}{*}{Math-FP16}
  & 15    & 5.76  & 7.13  & \textcolor{red!70!black}{$0.59\times$} \\
  & 30    & 17.68 & 18.72 & \textcolor{red!70!black}{$0.34\times$} \\
  & $>$30 & \multicolumn{3}{c}{\textcolor{red!70!black}{\textit{OOM}}} \\
\midrule
\multirow[c]{7}{*}{xFormers-FP32}
  & 15  & 3.37   & 6.26  & $1.00\times$ \\
  & 30  & 5.94   & 6.55  & $1.00\times$ \\
  & 100 & 23.94  & 11.50 & $1.00\times$ \\
  & 150 & 43.05  & 15.03 & $1.00\times$ \\
  & 200 & 65.81  & 18.56 & $1.00\times$ \\
  & 300 & 129.97 & 25.63 & $1.00\times$ \\
  & 400 & 218.30 & 32.70 & $1.00\times$ \\
\midrule

\multirow[c]{7}{*}{\textbf{ELSA-FP32}}
  & \cellcolor{red!4}15
  & \cellcolor{red!4}\textbf{3.58}
  & \cellcolor{red!4}\textbf{6.26}
  & \cellcolor{red!4}\textcolor{red!70!black}{$0.94\times$} \\
  & \cellcolor{red!4}30
  & \cellcolor{red!4}\textbf{6.66}
  & \cellcolor{red!4}\textbf{6.55}
  & \cellcolor{red!4}\textcolor{red!70!black}{$0.89\times$} \\
  & \cellcolor{gray!8}100
  & \cellcolor{gray!8}\textbf{20.22}
  & \cellcolor{gray!8}\textbf{11.50}
  & \cellcolor{gray!8}\textcolor{ForestGreen}{\textbf{$1.18\times$}} \\
  & \cellcolor{gray!13}150
  & \cellcolor{gray!13}\textbf{34.93}
  & \cellcolor{gray!13}\textbf{15.03}
  & \cellcolor{gray!13}\textcolor{ForestGreen}{\textbf{$1.23\times$}} \\
  & \cellcolor{gray!18}200
  & \cellcolor{gray!18}\textbf{50.76}
  & \cellcolor{gray!18}\textbf{18.56}
  & \cellcolor{gray!18}\textcolor{ForestGreen}{\textbf{$1.30\times$}} \\
  & \cellcolor{gray!23}300
  & \cellcolor{gray!23}\textbf{97.89}
  & \cellcolor{gray!23}\textbf{25.63}
  & \cellcolor{gray!23}\textcolor{ForestGreen}{\textbf{$1.33\times$}} \\
  & \cellcolor{gray!28}400
  & \cellcolor{gray!28}\textbf{158.04}
  & \cellcolor{gray!28}\textbf{32.70}
  & \cellcolor{gray!28}\textcolor{ForestGreen}{\textbf{$1.38\times$}} \\
\bottomrule

\end{tabular}
\caption{\textbf{FastVGGT scaling.} ELSA-FP32 sustains $1.18$–$1.38\times$
  speedup at long sequences with identical memory overhead.
  Math baselines fail beyond 30 frames (OOM); shading intensity reflects
  speedup magnitude, red denotes below-baseline performance.}
\label{tab:fastvggt}
    \vspace{-3mm}
\end{table}

\section{CUDA C++ Backend and Portability}
\label{sec:CUDA}

While our primary implementation uses Triton for rapid prototyping, we
provide a CUDA C++ backend for deployments without JIT support or
requiring finer control over the memory hierarchy.
Both backends share the same algorithmic core---a two-level prefix scan
over the monoid $(m,S,W)$---differing only in how that scan is realized:
within each block we apply a Hillis--Steele~\cite{hillis1986} or
Kogge--Stone~\cite{kogge1973} scan, and across blocks a
Blelloch scan~\cite{blelloch1990prefix} via cooperative groups and
CUB's \texttt{BlockScan} primitive~\cite{nvidia2024cub}, ensuring
portability across GPU generations.
Vectorization width and block size are exposed as template parameters and
autotuned at runtime.
Critically, the backend avoids Tensor-Core (HMMA/GMMA) instructions,
enabling FP32 execution on devices such as Jetson TX2 where Tensor Cores
are absent.
Preliminary internal benchmarks on A100, L4, and Jetson suggest that the C++ backend remains within roughly 10–15\% of the Triton implementation.

The C++ backend's fine-grained memory control proves particularly
advantageous under host-device offloading, where layer weights are
streamed from CPU memory one layer at a time.
Table~\ref{tab:offload_filled} reports results for
LLaMA-13B~\cite{touvron2023llama} (FP32, batch~1) using asynchronous
double-buffering across four sequence lengths.
At 4K and 8K tokens the workload is PCIe-bound: transfers complete faster
than ELSA's scan overhead, causing higher per-layer idle time (0.727\,ms
and 1.005\,ms vs.\ 0.228\,ms and 0.322\,ms for SDPA), and SDPA holds a
throughput edge ($-$42.4\% and $-$25.2\% for ELSA).
At $\ge$32K tokens, however, ELSA's lower memory footprint allows
computation to begin earlier, reducing per-layer wait from 540\,ms to
467\,ms and from 2069\,ms to 1695\,ms respectively, translating to
\textcolor{cvprblue}{\textbf{17.8\% and 20.2\% throughput gains}} over
SDPA, confirming that ELSA's memory efficiency compounds into end-to-end
gains as sequences and transfer times grow. Table~\ref{tab:llama8b} further reports LLaMA-8B direct inference
(no offloading), where ELSA's throughput gain scales from $+$1.7\%
at 4K to \textcolor{cvprblue}{\textbf{$+$12.3\%}} at 16K tokens.

\begin{table}[t]
    \centering
    \small
    \setlength{\tabcolsep}{4pt}
    \scalebox{0.7}{
    \begin{tabular}{llcccc}
        \toprule
        \textbf{Method} & \textbf{Metric} & \textbf{4K} & \textbf{8K} & \textbf{32K} & \textbf{65K} \\
        \midrule
        \multicolumn{6}{c}{\textit{FP32 offloading, LLaMA-13B, batch 1}} \\
        \midrule
        \multirow{2}{*}{ELSA}
            & Throughput (M\,tok/s)
                & 0.364
                & 0.956
                & \cellcolor{bestgray}\textbf{1.460}
                & \cellcolor{bestgray}\textbf{0.882} \\
            & Wait (ms/layer)
                & 0.727
                & 1.005
                & \cellcolor{bestgray}\textbf{467.2}
                & \cellcolor{bestgray}\textbf{1694.5} \\
        \midrule
        \multirow{2}{*}{SDPA}
            & Throughput (M\,tok/s)
                & \cellcolor{bestgray}\textbf{0.632}
                & \cellcolor{bestgray}\textbf{1.278}
                & 1.239
                & 0.734 \\
            & Wait (ms/layer)
                & \cellcolor{bestgray}\textbf{0.228}
                & \cellcolor{bestgray}\textbf{0.322}
                & 540.1
                & 2069.4 \\
        \midrule
        \multicolumn{2}{l}{Throughput gain}
                & \textcolor{red!70!black}{$-$42.4\%}
                & \textcolor{red!70!black}{$-$25.2\%}
                & \textcolor{ForestGreen}{\textbf{$+$17.8\%}}
                & \textcolor{ForestGreen}{\textbf{$+$20.2\%}} \\
        \bottomrule
    \end{tabular}}
    \caption{\textbf{LLaMA-13B host-device offloading.}
      At $\le$8K tokens the workload is PCIe-bound and ELSA's per-layer
      overhead exceeds its memory savings.
      At $\ge$32K tokens, ELSA's reduced footprint hides transfer latency
      more effectively, delivering \textbf{17.8--20.2\%} throughput gains.
      Gray shading marks the winner per metric and sequence length.}
    \label{tab:offload_filled}
    \vspace{-3mm}
\end{table}

\begin{table}[t]
\centering
\small
\setlength{\tabcolsep}{4pt}
\scalebox{0.74}{
\begin{tabular}{ccccc}
\toprule
\textbf{Sequence length} & \textbf{ELSA (tok/s)} & \textbf{SDPA (tok/s)} & \textbf{Gain} & \textbf{Latency $\downarrow$} \\
\midrule
4K  & 5{,}729.6 & 5{,}636.5 & \textcolor{ForestGreen}{$+$1.7\%}  & \textcolor{ForestGreen}{1.6\%} \\
\rowcolor{gray!10}
8K  & 4{,}862.9 & 4{,}440.6 & \textcolor{ForestGreen}{\textbf{$+$9.5\%}}  & \textcolor{ForestGreen}{\textbf{8.7\%}} \\
\rowcolor{gray!20}
16K & 3{,}636.4 & 3{,}237.7 & \textcolor{ForestGreen}{\textbf{$+$12.3\%}} & \textcolor{ForestGreen}{\textbf{11.0\%}} \\
\bottomrule
\end{tabular}}
\caption{\textbf{LLaMA-8B direct inference.}
  ELSA vs.\ ME-SDPA; gains scale with sequence length,
  reaching $+$12.3\% throughput and 11.0\% latency reduction at 16K tokens.}
\label{tab:llama8b}
\end{table}

%
\section{ELSA Variants}
\label{sec:Variants}

We examine ELSA's performance across different precision modes, long-context workloads, and
offload scenarios. Unless otherwise stated, throughput is reported in millions of tokens per
second (M\,tok/s) and memory refers to peak GPU usage.

\paragraph{FP32 variants and memory trade-offs.}
The main paper reports both strict FP32 and TF32-Turbo results. Strict FP32 is the primary headline for exact-attention claims; TF32-Turbo is an optional higher-throughput variant. We additionally benchmark a \textit{memory-lean} variant (streaming smaller key-value chunks controlled by $\eta$), which trades memory for throughput.
Table~\ref{tab:fp32variants} compares these variants against ME-SDPA at 16K tokens.

Strict FP32 achieves $3.4\times$ throughput improvement (1.36M vs.\ 0.40M tok/s) at a modest
memory increase from 0.289\,GB to 0.336\,GB. ELSA-Turbo (TF32) delivers an additional 5\%
speedup (1.43M tok/s) while preserving FP32-quality outputs; numerical error remains within
$10^{-4}$ relative to strict FP32. The memory-lean variant ($\eta{=}0.25$) matches
ME-SDPA's 0.288\,GB footprint while sustaining 0.13M tok/s. Taken together, these variants
span a continuum from maximum throughput to minimal memory, allowing practitioners to select
the operating point best suited to their hardware constraints.

\paragraph{FP16 regime and FlashAttention comparison.}
At 16K tokens in FP16, ELSA delivers 2.49M tok/s with 0.19\,GB memory.
FlashAttention-2~\cite{dao2023} achieves 2.88M tok/s
(${\sim}15\%$ faster) but requires 0.29\,GB;
FlashAttention-3~\cite{shah2024flashattention} reaches 2.73M tok/s at
0.36\,GB.
ME-SDPA sustains only 1.47M tok/s despite matching ELSA's 0.19\,GB
footprint.
Across sequence lengths, FA2 leads in raw throughput, but ELSA
consistently achieves the lowest memory usage among exact kernels and
narrows the speed gap as sequences grow.
ELSA thus carries its long-context advantages into FP16 while remaining
competitive with hardware-fused kernels.

\begin{table}[t]
    \centering
    \small
    \scalebox{0.8}{%
    \setlength{\tabcolsep}{4pt}
    \begin{tabular}{lcc}
        \toprule
        \textbf{Method} & \textbf{Throughput (M\,tok/s)} & \textbf{Memory (GB)} \\
        \midrule
        ME-SDPA                        & 0.40                                & 0.289 \\
        ELSA-Strict (FP32)               & 1.36                                & 0.336 \\
        ELSA-Turbo (TF32)                & \cellcolor{bestgray}{\textbf{1.43}} & \cellcolor{bestgray}{\textbf{0.288}} \\
        ELSA-Lean (FP32, $\eta{=}0.25$) & 0.13                                & \cellcolor{bestgray}{\textbf{0.288}} \\
        \bottomrule
    \end{tabular}}
    \caption{\textbf{FP32 variant comparison.}
      ELSA variants versus ME-SDPA at 16K tokens. Throughput in millions of tokens per
      second; memory is peak GPU usage. Strict FP32 (\texttt{allow\_tf32=0}) is used
      throughout the main paper; TF32-Turbo is an optional higher-throughput variant.}
    \label{tab:fp32variants}
    \vspace{-3mm}
\end{table}

\section{Limitations}
\label{sec:Limitations}

\begin{table}[t!]
\centering
\setlength{\tabcolsep}{4pt}
\resizebox{0.42\textwidth}{!}{
\begin{tabular}{c|cc|cc|c}
\hline
\multirow{2}{*}{\textbf{Batch size}} &
\multicolumn{2}{c|}{\textbf{Latency (ms)}} &
\multicolumn{2}{c|}{\textbf{Memory (GB)}} &
\multirow{2}{*}{\textbf{Speed-up}} \\
& \textbf{Math} & \cellcolor{gray!20}\textbf{ELSA} &
  \textbf{Math} & \cellcolor{gray!20}\textbf{ELSA} & \\
\hline

1  & {0.405} &
      \cellcolor{gray!20}{\textbf{0.399} \textcolor{ForestGreen}{(+1.5\%)}} &
      {0.0098} &
      \cellcolor{gray!20}{\textbf{0.0095} \textcolor{ForestGreen}{(+3.1\%)}} &
      1.02 \\

4  & {0.427} &
      \cellcolor{gray!20}{\textbf{0.424} \textcolor{ForestGreen}{(+0.7\%)}} &
      {0.0107} &
      \cellcolor{gray!20}{\textbf{0.0100} \textcolor{ForestGreen}{(+6.5\%)}} &
      1.01 \\

8  & {0.430} &
      \cellcolor{gray!20}{\textbf{0.411} \textcolor{ForestGreen}{(+4.4\%)}} &
      {0.0131} &
      \cellcolor{gray!20}{\textbf{0.0116} \textcolor{ForestGreen}{(+11.5\%)}} &
      1.05 \\

16 & {0.431} &
      \cellcolor{gray!20}{\textbf{0.409} \textcolor{ForestGreen}{(+5.1\%)}} &
      {0.0175} &
      \cellcolor{gray!20}{\textbf{0.0138} \textcolor{ForestGreen}{(+21.1\%)}} &
      1.05 \\
\hline
\end{tabular}}
\caption{\textbf{Swin-v2 window attention (FP16) across batch sizes.}
Latency and memory vs.\ Math baseline; ELSA columns shaded,
green denotes improvement.
ELSA consistently reduces both latency and memory with no retraining.}
\vspace{-3mm}
\label{tab:window}
\end{table}

\paragraph{Arithmetic Complexity.}
ELSA eliminates the $O(n^2)$ memory bottleneck but preserves quadratic
arithmetic complexity: the prefix-scan formulation parallelizes computation
and reduces memory traffic to $O(n)$, yet still performs all pairwise
interactions.
The term \emph{linear-scan} refers strictly to the additional memory and
I/O overhead, not to arithmetic complexity; ELSA is therefore best suited
to regimes where \emph{memory} rather than \emph{compute} is the primary
bottleneck.

\paragraph{Short-Sequence and Window Attention.}
For windowed MHSA (e.g., Swin with $n\!\le\!196$), the $O(\log n)$
scan-depth advantage diminishes and per-level merge overhead becomes
visible, while FA2/3 are highly optimized for small tiles.
Table~\ref{tab:window} shows that ELSA consistently outperforms the
unfused Math baseline in both latency and memory---improving by up to
5.1\% and 21.1\% respectively---yet FA2 remains faster in this
short-window FP16 regime.
This is consistent with ELSA's design goal: long/global contexts and
FP32 portability, where fused kernels provide no compatible alternative.

\paragraph{Training and Multi-GPU Scaling.}
The prefix-scan formulation introduces inter-block data dependencies that
complicate sequence parallelism: the monoid reduction across blocks must
complete before downstream layers proceed, creating synchronization
barriers that do not exist in standard attention.
Efficient multi-GPU scaling therefore requires careful pipeline scheduling
to overlap these reductions with communication, which remains future work.

\paragraph{Hardware Coverage.}
The current kernel is validated primarily on Ampere GPUs (A100, L4) and
Jetson TX2; broader coverage across Hopper, Ada Lovelace, and non-NVIDIA
accelerators is left to future work.